\documentclass[10pt,twocolumn,letterpaper]{article}

\usepackage[cvprfinal]{cvpr}
\usepackage{times}
\usepackage{epsfig}
\usepackage{graphicx}
\usepackage{enumitem}
\usepackage{amsmath}
\usepackage{amssymb}
\usepackage{stmaryrd} 
\usepackage{booktabs}
\usepackage{svg}

\usepackage{cvpr}
\definecolor{cvprblue}{rgb}{0.21,0.49,0.74}
\usepackage[pagebackref,breaklinks,colorlinks,citecolor=cvprblue]{hyperref}

\newcommand{\aka}{\textit{a.k.a.}}

\newcommand{\ours}{S3R-Net}
\newcommand{\approach}{unify-and-adapt}

\def\paperID{2} % *** Enter the Paper ID here
\def\confName{CVPR}
\def\confYear{2024}

\author{Nikolina Kubiak\\
University of Surrey\\
{\tt\small n.kubiak@surrey.ac.uk}
\and
Armin Mustafa\\
University of Surrey\\
{\tt\small armin.mustafa@surrey.ac.uk}
\and
Graeme Phillipson\\
BBC R\&D\\
{\tt\small graeme.phillipson@bbc.co.uk}
\and
Stephen Jolly\\
BBC R\&D\\
{\tt\small stephen.jolly@bbc.co.uk}
\and
Simon Hadfield\\
University of Surrey\\
{\tt\small s.hadfield@surrey.ac.uk}
}

\begin{document}

\title{S3R-Net: A Single-Stage Approach to Self-Supervised Shadow Removal} % Replace with your title

\maketitle
\begin{abstract}
\vspace{-0.1cm}
In this paper we present \textit{\ours}, the Self-Supervised Shadow Removal Network. The two-branch WGAN model achieves self-supervision\footnote {In this work, we take "self-supervised" to mean a system that does not require ground truth shadow-free pairs for its training inputs, but only a set of shadow-free data from unrelated scenes. This matches the supervision requirements of existing self-supervised techniques, \eg\ cycleGANs.} relying on the \textit{\approach} phenomenon - it unifies the style of the output data and infers its characteristics from a database of unaligned shadow-free reference images. This approach stands in contrast to the large body of supervised frameworks. \ours\ also differentiates itself from the few existing self-supervised models operating in a cycle-consistent manner, as it is a non-cyclic, unidirectional solution. The proposed framework achieves comparable numerical scores to recent self-supervised shadow removal models while exhibiting superior qualitative performance and keeping the computational cost low. The pre-trained models and the code can be found in \href{https://github.com/n-kubiak/S3R-Net}{our github repo}. \looseness-1
\end{abstract}

\vspace{-0.2cm}
\section{Introduction}
Shadows are physical phenomena that arise when an obstacle appears on the trajectory between a light source and a surface. They can provide necessary guidance for three-dimensional understanding of objects \cite{barron_2014_shape} and scenes \cite{eigen_2014_depth,zhang_1999_sfs} in monocular settings \cite{howard_2012_depth}. However, in other scenarios, they can negatively affect our perception of the world around us. Therefore, a number of shadow removal frameworks have been developed for aesthetic purposes, ranging from de-shadowing casual capture photos \cite{zhang_portrait_2020} to cropping out unwanted objects alongside their shadows \cite{lu_2021_omnimatte,zhang_2021_noshadow}. Shadows can also have an adverse affect on the functioning of automated systems. In particular, they can obscure key visual clues, or introduce ambiguities by resembling dark objects \cite{wu_2012_shadowsdriving}. Thus, shadow removal models can be useful as a pre-processing step for other computer vision tasks, \eg\ for document shadow removal \cite{georgiadis2023lp,jung_2018_documents} or for improving the accuracy of autonomous vehicle systems \cite{wang_2020_vehicles}.

Unfortunately, most of the existing shadow removal frameworks operate in a supervised manner. The models require paired shadowed and shadow-free images, and may also required matching shadow masks. Due to the changing nature of the sun and sky conditions, such ground truth data is difficult to capture consistently for outdoor scenes \cite{qu_deshadownet_2017,wang_stacked_2018}. Indoor scenarios, while easier, have not gathered significant interest. Finally, synthetic datasets, \eg\ \cite{sidorov_2019_conditional}, offer perfect colour and pixel-wise alignment yet they lack the realism and detail of data captured in the wild. 

It is obviously possible to limit the training requirements of the shadow removal models and, in fact, a small number of such methods have been published. In the past few years, a few self-supervised methods \cite{hu_mask-shadowgan_2019,jin_2021_unsup,liu_shadow_2021,Vasluianu_2021_CVPR} have exploited cycle-consistency as their main supervisory signal. However, cycleGAN-based models require a secondary proxy task (here: shadow generation). This can negatively affect the model's robustness as it relies on both tasks functioning correctly and in balance. Additionally, this approach increases the number of generators and discriminators used by the system and, thus, the model's complexity. To the best of our knowledge, only a single unsupervised shadow removal framework has been proposed \cite{he_2021_unsupervised} yet its high domain-specificity limits its breadth of applications. 

\begin{figure*}[h]
\begin{center}
\includegraphics[width=\textwidth]{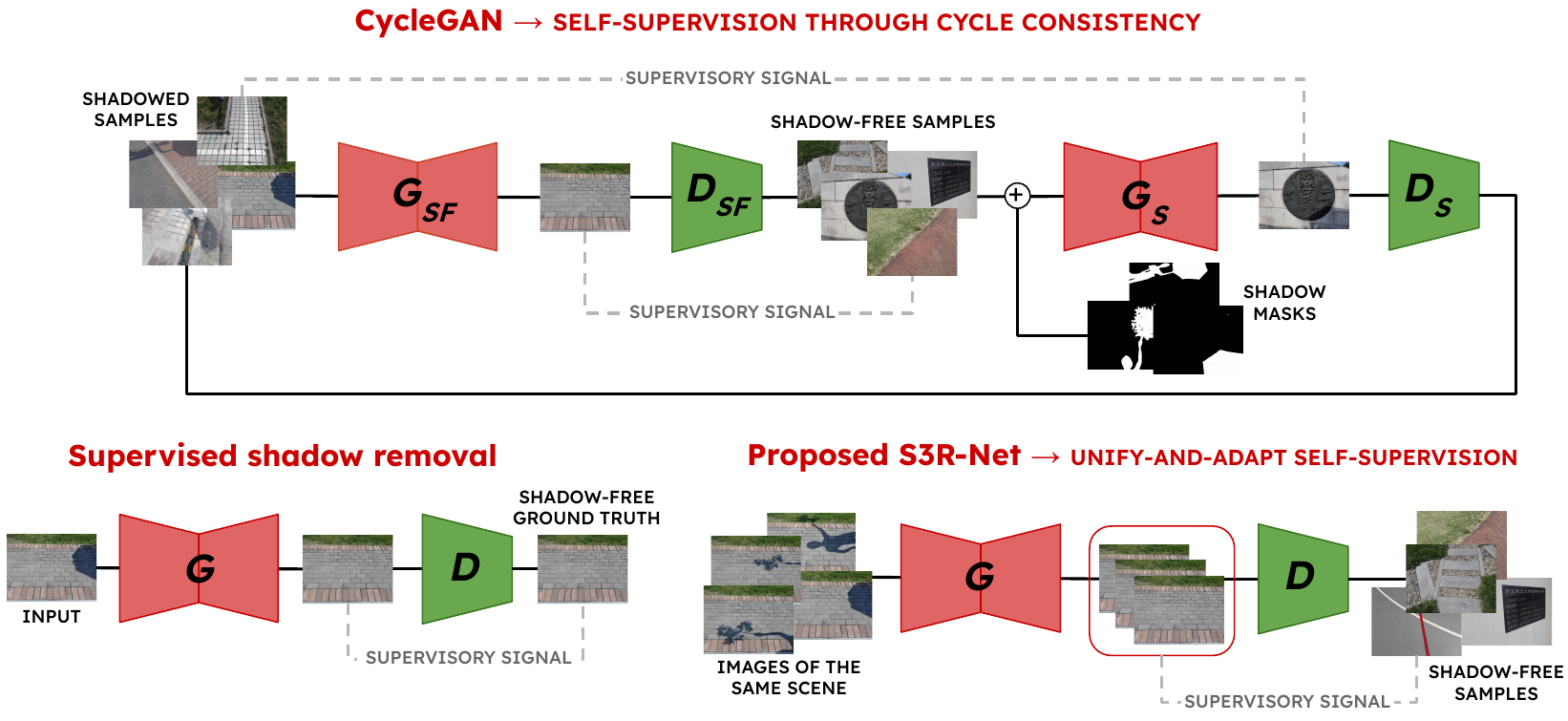}
\end{center}
\caption{We present the architectures of a standard supervised shadow removal model (bottom left), a cycle-consistent self-supervised model (top) and the proposed \ours\ (bottom right), exploiting a \approach\ approach to self-supervision. The figure shows model inputs, key modules and the sources of supervisory signal. G and D denote the generator and the discriminator of a GAN, and SF/S subscripts are used to indicate networks generating/discriminating shadow-free/shadowed data. }
\label{comparison}
\vspace{-0.2cm}
\end{figure*}

Motivated by the above, we present a novel solution with significantly reduced supervision requirements. The proposed Self-Supervised Shadow Removal Network (\textit{\ours}), contrasted with existing architectures in Fig.\ \ref{comparison}, is trained without the need for paired shadowed and shadow-free images. The model does not require any ground truth shadow masks nor does it rely on the accuracy of any explicit shadow detection modules. Instead, the desired appearance is learnt via adversarial learning from a collection of shadow-free images. This reference database does not have to be paired or aligned with the input sequences, and can represent any scene, including scenes which do not appear in the shadowed training data. The driving force of our GAN framework is the \textit{\approach} approach. The de-shadowed outputs are created by a unidirectional, two-branch network that attempts to map multiple differently shadowed versions of a scene to a uniform shadow-free output (the \textit{unify} step). The style of this unified output domain is then adapted to that of the reference style via a discriminator which distinguishes between the generated and the real shadow-free samples (the \textit{adapt} step). This approach helps us ensure good colour-consistency and overall quality of the reconstructions while exploiting both cross-scene and self-consistency information.%We also note that at inference time the system has only one branch.

To sum up, the contributions of this paper are as follows:
\begin{enumerate}
    \item We present a new \textit{\approach} self-supervised shadow removal model that achieves competitive scores on the ISTD and AISTD datasets without relying on cycle-consistency or domain-specific priors; \looseness-1
    % \vspace{-0.1cm}
    \item We demonstrate that \ours\ achieves superior qualitative performance when contrasted with the best performing and most recent self-supervised shadow removal frameworks;
    % \vspace{-0.1cm}
    \item We prove the efficiency of the proposed system via a model parameter count and train-time GFLOPS comparisons between existing self-supervised shadow removal models (see: Fig.\ \ref{flops}).
    \vspace{-0.1cm}
\end{enumerate}

\section{Literature review}
Shadow removal is a not a new computer vision problem. Historically, the literature in the field looked at colour and illumination statistics to create physics-based solutions, \eg\ \cite{drew_2003_recovery,finlayson_2001_cameracalib,shor_2008_shadow}. Other works relied on user input to guide the shadow detection and removal steps \cite{gong_2016_interactive,gryka_2015_softshadows}. However, in recent years, we have observed the emergence of large-scale shadow removal datasets, coupled with the rising popularity of deep learning. These changes have led to the creation of a number of learning-based solutions that have produced state-of-the-art results in shadow removal and its sister task of shadow detection \cite{vicente_2015_detection,wang_2021_det,wang_2020_shadowdet,wang_2021_bmvc,zhang_2019_det}. In this literature review, we will focus on the learnt de-shadowing frameworks and, in particular, their approach to reducing supervision requirements.

\textbf{Learnt shadow removal. }
% \vspace{-0.2cm}
A number of works draw inspiration from physical illumination models that find the mapping between shadowed and shadow-free pixels \cite{fu_2021_auto,le_2019_via}. Such a function is used to over-expose the shadowed data so that its dark areas match the shadow-free regions. Then, the original and over-exposed images are blended to achieve a de-shadowed result. %Vicente \etal\ \cite{vicente_2017_detremoval} achieve this using a histogram. 
SP-Net \cite{le_2019_via} learns the shadow parameters and uses them to combine the natural and over-exposed shadowed data using a matte. Fu \etal\ \cite{fu_2021_auto} formulate the task as an auto-exposure fusion problem and smartly weigh a number of over-exposed shadowed regions to de-shadow the input.

%Another group of papers exploits the fact that CNN-based classifiers trained on ImageNet can generalise well across different datasets and tasks depending on semantic scene understanding. 
Another group of papers exploits semantic scene understanding contained in pre-trained backbones. DeShadowNet \cite{qu_deshadownet_2017} uses features from shallower and deeper layers of the VGG classifier \cite{simonyan_2015_vgg} to decode appearance and semantic scene information that can be combined to guide shadow removal. Cun \etal\ \cite{cun_2020_HAN} fuse the features with the input and use hierarchical feature aggregation to combine spatial attention with the information from earlier layers. Hu \etal\ \cite{hu_2020_context} use CNN features to learn the direction-aware spatial context used to guide shadow removal. CANet \cite{chen_2021_canet} matches contextual patches between shadow-free and shadowed regions, and transfers the features at different scales from the former to the latter. In DeS3 \cite{jin2022des3} vision transformer features are used alongside attention and colour constancy constraints. PRNet \cite{wang2024progressive} links the features with RNNs.

Generative Adversarial Networks (GANs) have also been a common choice for shadow removal. Wang \etal\ \cite{wang_stacked_2018} stacked two GANs to detect the shadow and then use its mask as conditioning information for the removal step. RIS-GAN \cite{zhang_2020_risgan} consists of three parallel GANs for shadow removal as well as residual and illumination estimation. ARGAN \cite{ding_argan_2019} uses attention to recursively detect and remove shadows, making it robust to shadows of varying strength and complexity. SHARDS \cite{sen2023shards} deshadows low-resolution images and then uses them as guidance for full-resolution shadow removal. More recently, solutions using state-of-the-art techniques such as transformers \cite{guo2023shadowformer,yu2023cnsnet} or diffusion models \cite{guo2023shadowdiffusion,liu2024recasting,mei2024latent} have also been proposed. 

Some authors have also explored new ways of thinking about shadow removal. Li \etal\ \cite{li2023leveraging} demonstrated that the task of inpainting is compatible with shadow removal, and linking the two decreases the prominence of shadow remnants. The authors also propose a system \cite{li2023learning} relying on attentional fusion of 2 task branches -- one for shadow-free region information relay and one for deshadowing. Wan \etal\ \cite{wan2022style}, on the other hand, pose shadow removal as a intra-frame style transfer problem.

Unlike the above, some systems do not require ground truth shadow-free images. Le and Samaras \cite{le_2020_shadow} build on SP-Net \cite{le_2019_via} and require only paired shadow masks to train their weakly-supervised model. These are used to crop out partly-shadowed and shadow-free patches from an image and limit the dependence on paired shadow-free data. In contrast, Liu \etal\ \cite{liu_2021_shadow_gen} create a train set by masking out shadowed and shadow-free areas. Their G2R-ShadowNet generates shadows, guided by the real shadow area masks, and then learns to de-shadow them using the shadow-free input regions. Zhong \etal\ \cite{zhong2022shadow} expand on this solution by improving the realism of the generated shadows. Guo \etal\ \cite{guo2023boundary} use masks to train an image decomposition module being part of their diffusion-based shadow removal system.

The shadow removal literature is rich yet all of the above methods rely on ground truth in the form of shadow-free images and/or shadow masks. To improve generality, it is important for models to be trained on large-scale, real-world datasets. Unfortunately, capturing ground truth shadow-free data is error-prone and time-consuming. This is apparent in the existing shadow datasets \cite{qu_deshadownet_2017,wang_stacked_2018} which are known to contain slight scene framing and colour inconsistencies between different images corresponding to the same scene \cite{hu_2020_context,le_2019_via,Vasluianu_2021_CVPR}. Obtaining the masks for real-life data is also laborious as it requires per-pixel annotations. 

\textbf{Un- and self-supervised frameworks}
Motivated by the discussed challenges, this paper aims to create a shadow removal network that does not require aligned ground truth data. Instead of relying solely on pixel-wise error computation between the input and the reference, the system should guide the shadow removal using adversarial losses and exploit other information present in the data. %Similar objectives have been set by a small number of authors whose works we review in the following paragraphs.

A common approach to self-supervised learning relies on cycleGANs \cite{CycleGAN_2017} and cycle-consistency training. The images used in this bidirectional process still come from two domains -- shadowed and shadow-free -- yet they no longer have to be paired, which lowers the data requirements. Mask-ShadowGAN \cite{hu_mask-shadowgan_2019} is the first self-supervised shadow removal framework and it operates based on the generic cycleGAN losses. Liu \etal\ build on this solution in LG-ShadowNet \cite{liu_shadow_2021} and introduce two key changes: they first learn shadow removal in the L channel of the Lab colour space and then warmstart the all-channel network with those weights, and propose a vector-based colour loss. Vasluianu \etal\ \cite{Vasluianu_2021_CVPR} focus on the colour and pixel-wise inaccuracies in the existing datasets. To tackle them, they blur the inputs and outputs for colour-consistency enforcement, and rely purely on perceptual losses to control the content and style. DC-ShadowNet \cite{jin_2021_unsup} expands the cycleGAN idea to soft and hard shadow understanding, and uses shadow domain classifiers alongside physics-based constraints.

Cycle-consistency is not the only way of reducing the supervision requirements. He \etal\ \cite{he_2021_unsupervised} propose an unsupervised portrait-specific solution that uses GAN inversion and leverages the generative facial priors embedded in the pre-trained StyleGAN2 \cite{stylegan2}. While such priors are readily available for the popular portrait domain, the method cannot be easily interpolated to other more arbitrary problems.  

In contrast, this paper follows a multi-branch \textit{\approach} approach to self-supervision. This avoids the dependency on a paired inverse task, and instead exploits the commonality of de-shadowed outputs across different shadowed training inputs. This removes the need to train an inverse task, improving compactness, efficiency and robustness of the system.

\section{Methodology}
In the following paragraphs we discuss the implementation details of the proposed method. First, we explain our approach to self-supervision. This is followed by a discussion of the losses driving our model. Finally, we provide more details on the implementation of the GAN used as part of our model. 
\begin{figure*}[t]
\begin{center}
\includegraphics[width=15cm]{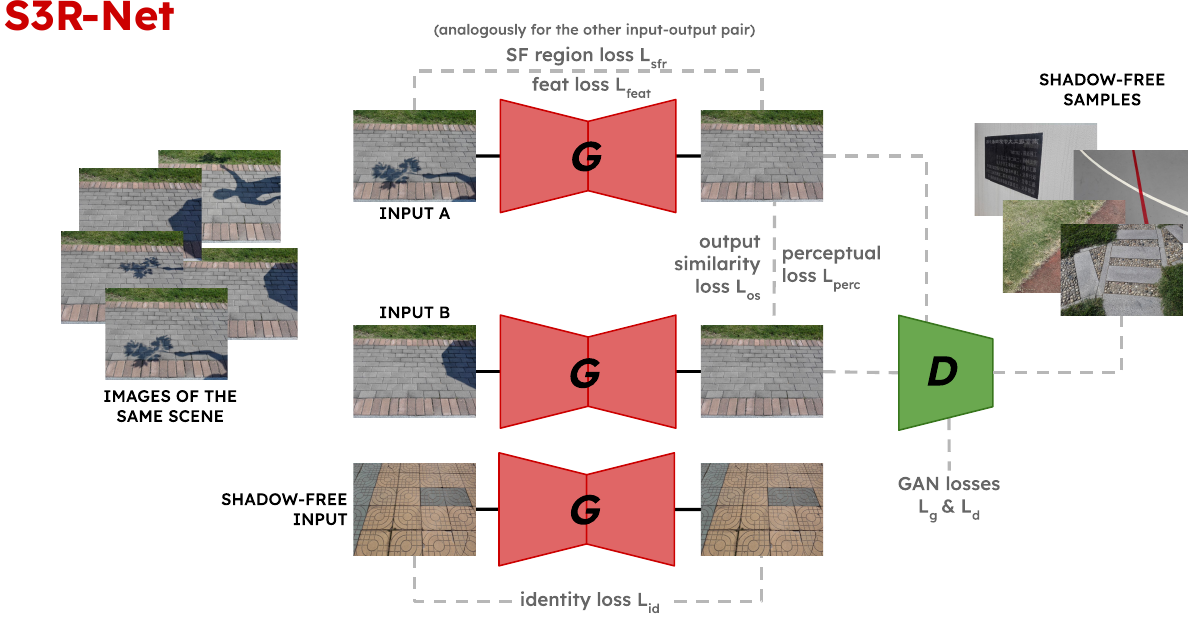}
\end{center}
\vspace{-0.1cm}
\caption{Our \ours\ system and its losses. The generators (G) shown above are the same exact model with the same weights. }
\vspace{-0.1cm}
\label{model}
\end{figure*}

\subsection{The \approach\ approach to self-supervision and other training losses} 
\label{training_losses}
To limit the need for ground truth, we propose a self-supervised solution. The few existing domain-independent models that do not require paired data to train, all operate in a cycle-consistent (\ie\ bidirectional) manner. In contrast, we build \ours\ based on an emerging \textit{\approach} approach to self-supervision from the field of relighting, which uses a unidirectional, single-stage 2-branch network \cite{kubiak_2021_silt} (Fig.\ \ref{model}). In such a GAN-based architecture (described in more detail in Section \ref{GAN}) the input (shadowed) images \textbf{I} corresponding to the same scene are paired and fed into the generator $\mathcal{G}$ in parallel. The generator produces a shadow-correction residual that is added to the input before going through the final activation layer. The final reconstruction can therefore be described as $\hat{\textbf{I}} = \textbf{I} + \mathcal{G}(\textbf{I})$. Using residuals is intended to limit the region of change within the image and keep the shadow-free areas intact. Instead of relying on a pixel-wise aligned ground truth, the system exploits the knowledge that the correct de-shadowed solution must be consistent across all differently shadowed versions of the input scene. Once the generator has learnt to enforce uniformity across the different variations of the input, the discriminator $\mathcal{D}$ helps to \textit{adapt} this uniform output towards the correct output style through its adversarial losses (Eq.\ \ref{gan_loss_g}-\ref{gan_loss_d}). Here, this target style is inferred from a collection of shadow-free samples $\textbf{I}^*$ showing arbitrary scenes. 

%To enforce the similarity between the outputs, we introduce the output similarity loss
In addition to the control achieved through $\mathcal{D}$, we exploit a number of generator losses. As discussed in the Introduction, the available datasets have colour- and pixel-wise misalignments between the inputs and shadow-free equivalents of the same scene. With this in mind, we want to control our training using a mixture of pixel-wise and feature-based losses. The former provide strong guidance signals yet are prone to innate dataset errors. The latter are not as strong yet are more resilient to pixel-wise discrepancies.

The first pair of proposed losses controls the \textit{unify} aspect of the \textit{\approach} approach, \ie\ ensures that both outputs of our 2-branch network look the same, regardless of the initial shadows present. We enforce this using an L1 loss $\mathcal{L}_{os}$ defined as
\begin{equation}
    \vspace{-0.05cm}
    \mathcal{L}_{os} =  \lVert (\textbf{I}_A + \mathcal{G}\left(\textbf{I}_A\right)) - (\textbf{I}_B + \mathcal{G}\left(\textbf{I}_B\right)) \rVert_1. \\%\times \left(\frac{1}{4}\right)^i \\
    %\quad and \quad x \in \left(1,\frac{1}{2},\frac{1}{4}\right).
    \label{os_loss}
    \vspace{-0.05cm}
\end{equation}
In the above equation, we use the $A$ and $B$ subscripts to refer to the images associated with each of the two network branches. It is also important to note that the output similarity loss is not applied directly to the shadow-correction residuals (\ie\ $\mathcal{G}(\textbf{I})$), but rather to the re-composited de-shadowed images. 

The uniformity of the outputs is additionally controlled using the perceptual loss $\mathcal{L}_{perc}$. This compares the features extracted from both outputs using a pretrained VGG-19 backbone $vgg$. The loss can be formalised as
\begin{equation}
\vspace{-0.05cm}
    \mathcal{L}_{perc} =  \sum^i \Big\lVert vgg_i\left(\hat{\textbf{I}}_A\right) - vgg_i\left(\hat{\textbf{I}}_B\right) \Big\rVert_1 \\
    \label{perc_loss}
    \vspace{-0.05cm}
\end{equation}
and \textit{i} represents different feature scales within the network.

While $\mathcal{L}_{os}$ and $\mathcal{L}_{perc}$ aim to equate the outputs, any colour or framing differences between the paired images will affect the output quality. We can counteract this by preserving the information present in the shadow-free regions. To this end, we calculate a shadow mask $\textbf{M}$ for each branch's input-output pair by applying Otsu tresholding \cite{otsu_1979_threshold} to a greyscale version of the images, \ie\ $\textbf{M} = Otsu(\textbf{I}-\hat{\textbf{I}})$. In the mask, 1s denote shadowed regions and 0s - shadow-free. To focus on the shadow-free areas, we invert the mask and obtain $\hat{\textbf{M}}$. We then use $\hat{\textbf{M}}$ to mask out the shadowed region in the input and the output, and compare the visible shadow-free areas. The resulting shadow-free region loss $\mathcal{L}_{sfr}$ is given as
\begin{equation}
\vspace{-0.05cm}
    \begin{aligned}
        \mathcal{L}_{sfr} = \Big\lVert \left(\hat{\textbf{M}}_A \odot \hat{\textbf{I}}_A\right) - \left(\hat{\textbf{M}}_A \odot \textbf{I}_A\right) \Big\rVert_2 + \\
        \Big\lVert \left(\hat{\textbf{M}}_B \odot \hat{\textbf{I}}_B\right) - \left(\hat{\textbf{M}}_B \odot \textbf{I}_B\right) \Big\rVert_2,
        \label{sfr_loss}
    \end{aligned}
    \vspace{-0.05cm}
\end{equation}
where $\odot$ symbolises the Hadamard product.

Even though comparing information from the output and its corresponding input is robust against pixel-wise discrepancies, there is potential for error stemming from imperfect mask calculation. Therefore, we also add a feature-based counterpart to $\mathcal{L}_{sfr}$ -- a feature loss $\mathcal{L}_{feat}$, originally proposed in \cite{jin_2021_unsup}. In their paper, Jin \etal\ conduct a study on features extracted from shadowed and shadow-free images of the same scene, and discovered that the features extracted at a particular network layer (Conv22 in the pretrained VGG-16 backbone -- ${vgg_{22}}$) are the most shadow-invariant. Thus, we can extract features from the model input and the produced output, and ensure their feature-level consistency regardless of shadows. The aforementioned loss can be described as
\begin{equation}
\begin{aligned}
\vspace{-0.05cm}
    \mathcal{L}_{feat} =  \Big\lVert vgg_{22}\left(\hat{\textbf{I}}_A\right) - vgg_{22}\left(\textbf{I}_A\right) \Big\rVert_1 +  \\
    \Big\lVert vgg_{22}\left(\hat{\textbf{I}}_B\right) - vgg_{22}\left(\textbf{I}_B\right) \Big\rVert_1 .
    \label{feat_loss}
    \vspace{-0.05cm}
\end{aligned}
\end{equation}
While $\mathcal{L}_{sfr}$ is focused on the actual pixel values, $\mathcal{L}_{feat}$ is more concerned with general scene structure preservation, and the losses complement each other.

Finally, we want to prevent the shadow removal model from uniformly brightening the entire image. Therefore, we add a constraint that makes $\mathcal{G}$ de-shadow only the shadowed regions and leave the shadow-free areas intact. % In would be undesirable for \ours\ to lighten up all pixels equally as this could skew the colour and brightness in the image. 
 The shadow-free region loss goes some way towards enforcing this. However, in the absence of ground truth shadow masks, $\mathcal{L}_{sfr}$ must rely on the network output and the calculated shadow masks $\hat{\textbf{M}}$, which limits its robustness. We mitigate this by feeding $\mathcal{G}$ a shadow-free image $\textbf{I}_{sf}$ and expecting it to produce a virtually empty residual, leading to a no-adjustment reconstruction $\hat{\textbf{I}}_{sf}$. We enforce this using the identity loss
\begin{equation}
    \mathcal{L}_{id} = \big\lVert \textbf{I}_{sf} -  \hat{\textbf{I}}_{sf} \big\rVert_1.
    \label{id_loss}
\end{equation}
We would like to emphasise that $\textbf{I}_{sf}$ never represents the same scene as the $\{\textbf{I}_A, \textbf{I}_B\}$ input pair.

Since the losses described above serve different purposes and address different problems, we add scaling $\lambda$ to each sub-loss. The total generator loss thus becomes
\begin{equation}
\begin{aligned}
    \mathcal{L}_{total} = \mathcal{L}_{g} + \lambda_{os}\mathcal{L}_{os}  + 
 \lambda_{perc}\mathcal{L}_{perc}  + \\
 \lambda_{sfr}\mathcal{L}_{sfr} + \lambda_{feat}\mathcal{L}_{feat}  + \lambda_{id}\mathcal{L}_{id}.
    \label{total_loss}
\end{aligned}    
\end{equation}

The 2-branch approach described in this section is necessary for loss calculation during training. However, at test time, the losses are no longer used and, thus, branch duplication is not required. Consequently, this means that during inference, we only need to feed a single image into a single $\mathcal{G}$ branch and there is no need to pair up the inputs.

\subsection{The adversarial shadow removal model}
\label{GAN}
As outlined above, the core of the proposed \ours, shown in Fig.\ \ref{model}, is a GAN.
Our system builds on the classic pix2pixHD model \cite{wang_2018_pix2pixHD}. The framework uses a fully-convolutional encoder-decoder network as its generator and has a fully-convolutional multi-scale discriminator.

We train the aforementioned GAN as a Wasserstein GAN (WGAN) \cite{arjovsky_2017_wgan}. This approach brings in a few noteworthy changes. WGANs abandon the 0-1 (fake-real) labels of vanilla GANs \cite{goodfellow_2014_gan}. Instead, their underlying loss metric, the Wasserstein distance $\mathbb{W}$ (\aka\ Earth-mover's distance), 
\begin{equation}
\vspace{-0.05cm}
    \mathbb{W}(\textbf{I},\textbf{I}^*) = \mathbb{E}[\mathcal{D}\left(\mathcal{G}\left(\textbf{I}\right)\right)] - \mathbb{E}[\mathcal{D}\left(\textbf{I}^*\right)]
    \vspace{-0.05cm}
\end{equation}
can be understood as the minimum cost of aligning all of one distribution's samples with the other's. 

\begin{figure*}[t]
\begin{center}
\footnotesize
\begin{tabular}{ c @{\hspace{0.05cm}} c @{\hspace{0.05cm}} c @{\hspace{0.05cm}} c @{\hspace{0.05cm}} c @{\hspace{0.05cm}} c @{\hspace{0.05cm}} c @{\hspace{0.05cm}} c }
Input & base &base + $\mathcal{L}_{sfr}$ & base + $\mathcal{L}_{id}$ & base + $\mathcal{L}_{sfr}$ + $\mathcal{L}_{id}$ &  + $\mathcal{L}_{perc}$ + $\mathcal{L}_{feat}$ & Reference \\
\includegraphics[width=2.4cm]{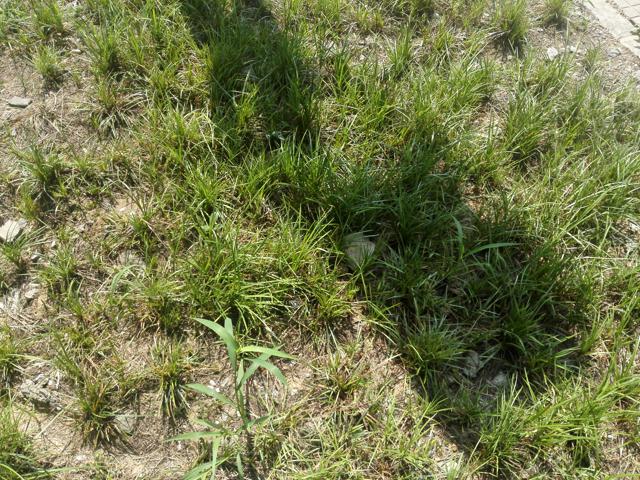}&
\includegraphics[width=2.4cm]{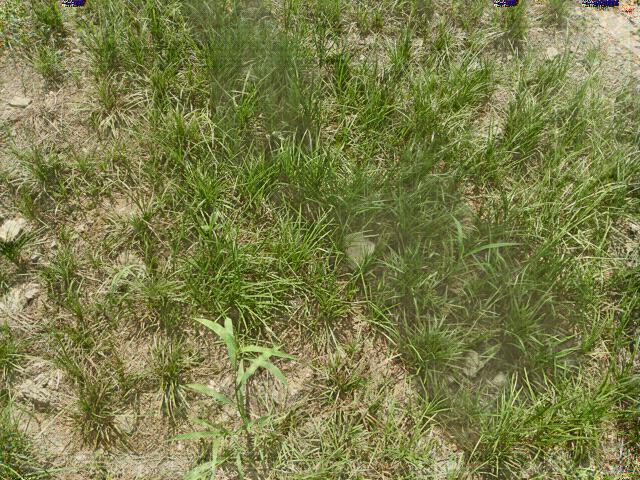}&
\includegraphics[width=2.4cm]{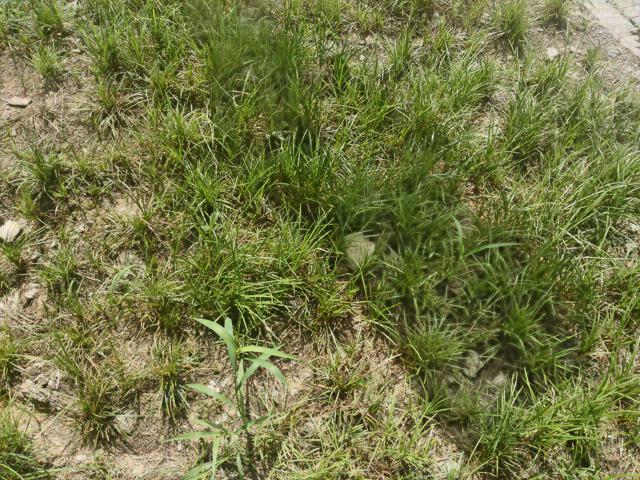}&
\includegraphics[width=2.4cm]{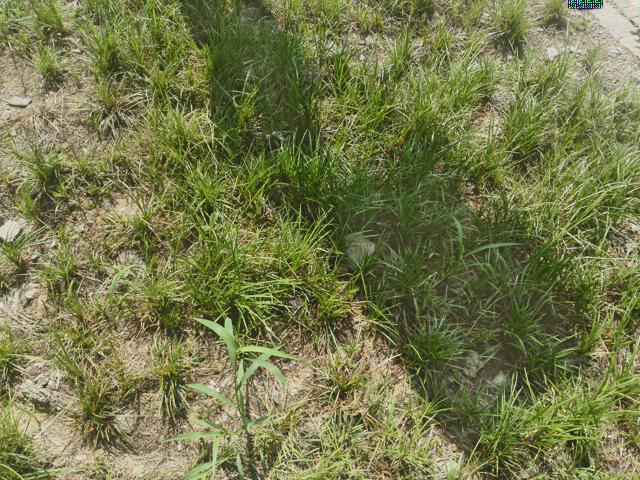}&
\includegraphics[width=2.4cm]{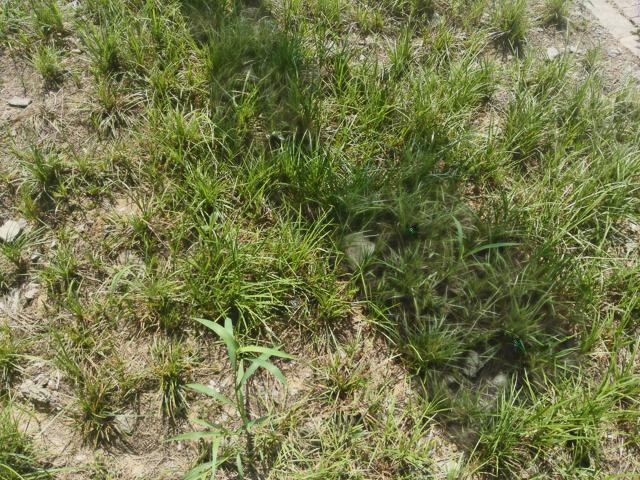}&
\includegraphics[width=2.4cm]{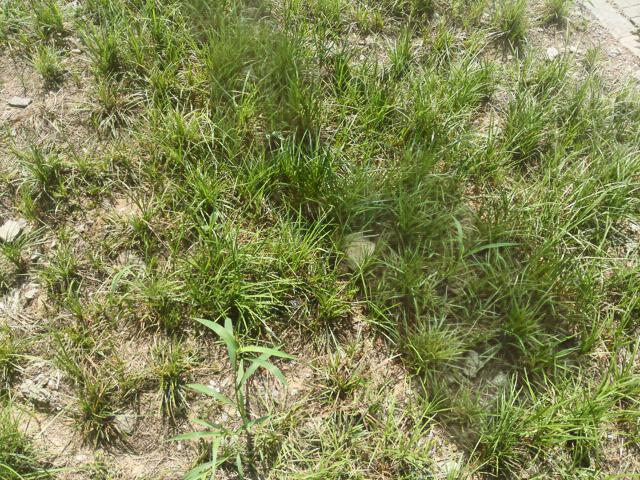}&
\includegraphics[width=2.4cm]{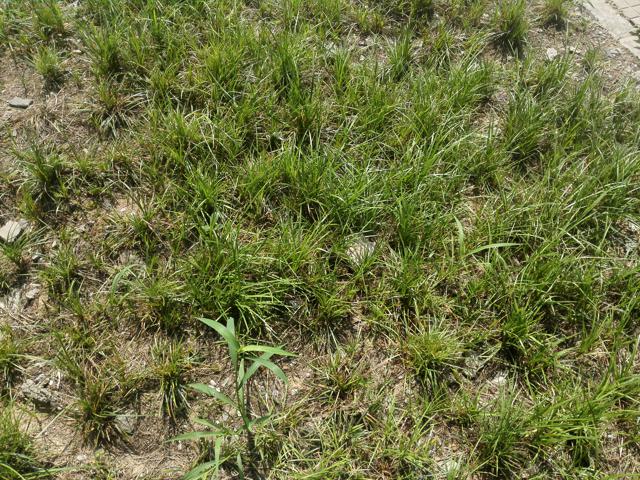}\\

\includegraphics[width=2.4cm]{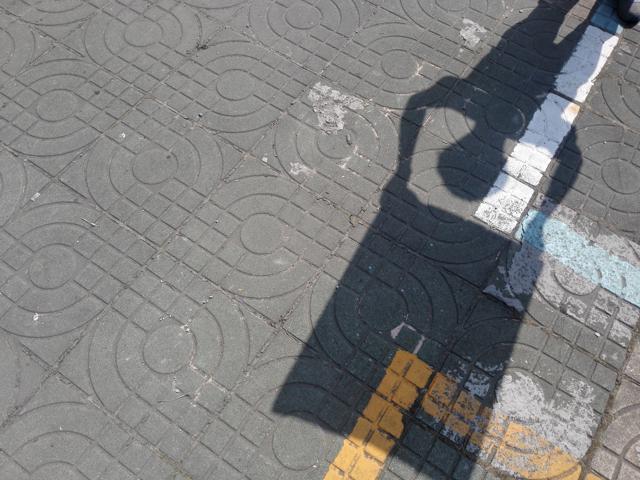}&
\includegraphics[width=2.4cm]{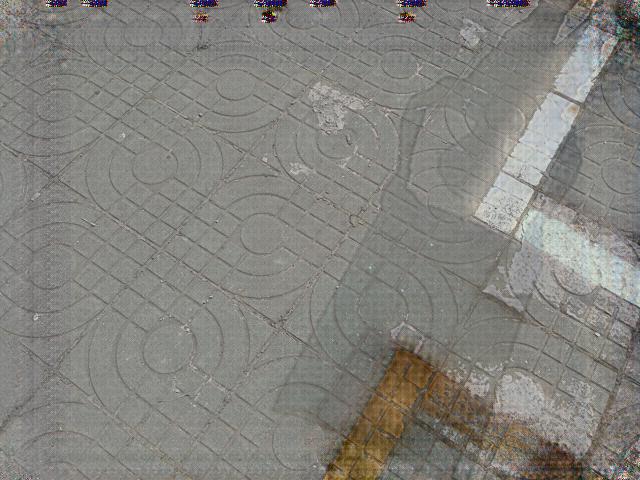}&
\includegraphics[width=2.4cm]{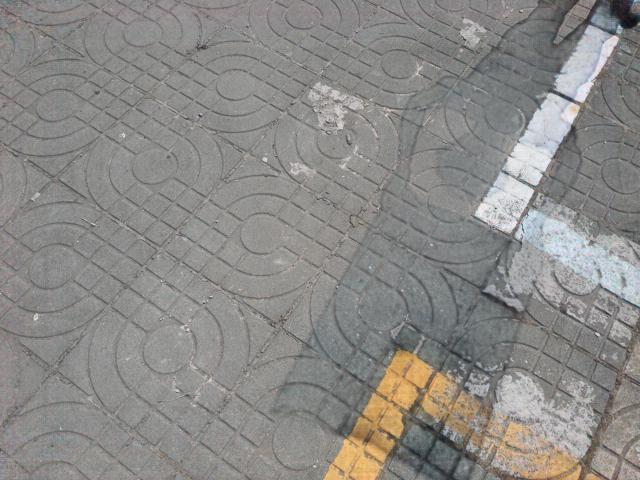}&
\includegraphics[width=2.4cm]{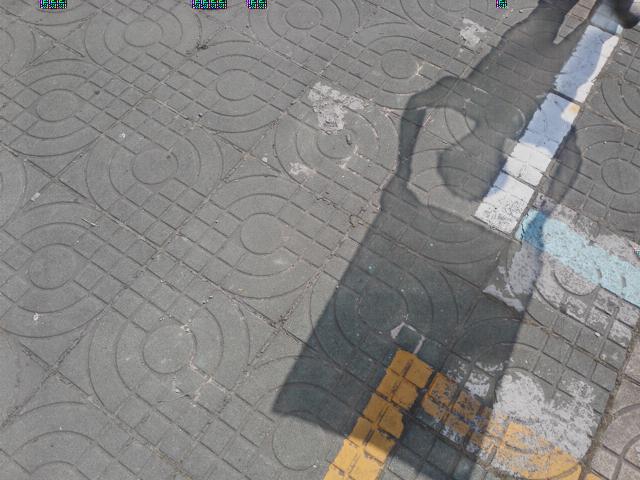}&
\includegraphics[width=2.4cm]{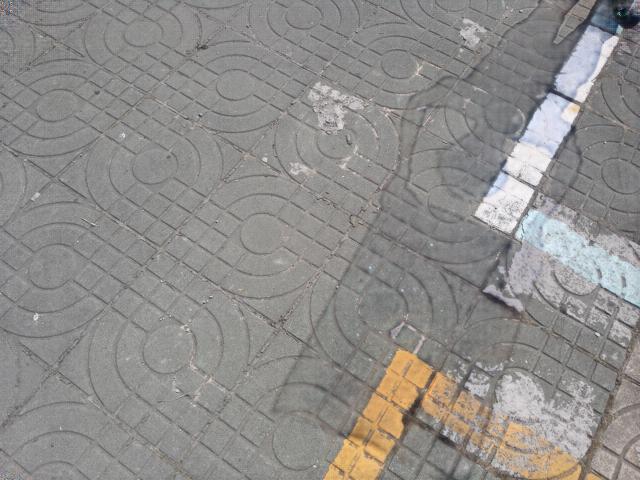}&
\includegraphics[width=2.4cm]{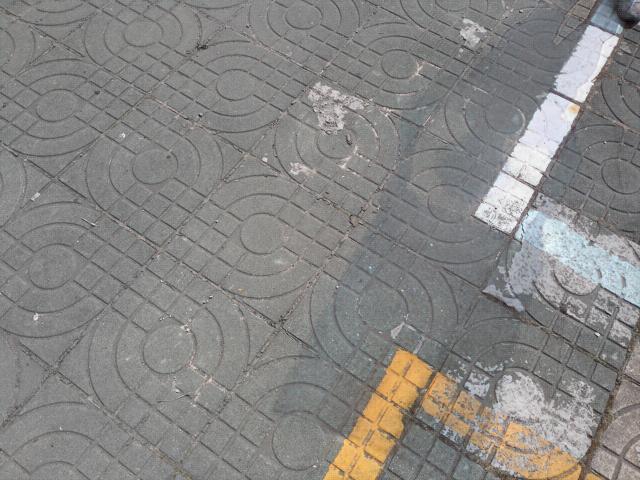}&
\includegraphics[width=2.4cm]{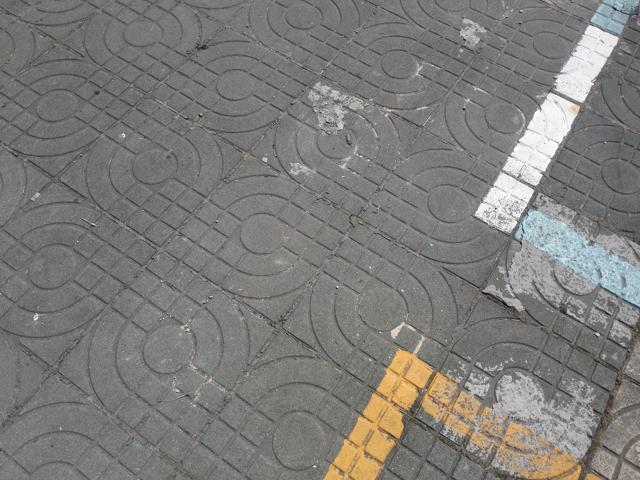}\\
\end{tabular}
\end{center}
\vspace{-0.3cm}
\caption{Ablation study: visual impact of \ours's losses. }
\vspace{-0.3cm}
\label{loss_ablation_fig}
\end{figure*}

To ensure the desired behaviour, we enforce 1-Lipschitz continuity on the discriminator function using a gradient penalty $\mathcal{E}_{gp}$ \cite{gulrajani_2017_wgan_gp}. To calculate $\mathcal{E}_{gp}$, the discriminator is fed an image $\bar{\textbf{I}}$ created by mixing real and generated examples with a weight sampled from a uniform distribution, 
\begin{equation}
\vspace{-0.05cm}
    \bar{\textbf{I}} = \epsilon \textbf{I} + (1-\epsilon)\textbf{I}^* \quad\mathrm{where}\quad \epsilon \sim \mathcal{U}(0,1).
    \vspace{-0.05cm}
\end{equation}
Then, the gradients w.r.t.\ the sample $\nabla_{\bar{\textbf{I}}}$ are constrained to be close to 1, \ie\ \begin{equation}
    \vspace{-0.05cm}
   \mathcal{E}_{gp} = \mathbb{E}[\left(\big\lVert\nabla_{\bar{\textbf{I}}}\mathcal{D}\left(\bar{\textbf{I}}\right)\big\rVert_2 - 1\right)^2].
    \label{gp}
    \vspace{-0.05cm}
\end{equation}
With all of this in mind, the adversarial $\mathcal{G}$ and $\mathcal{D}$ losses - $\mathcal{L}_{g}$ and $\mathcal{L}_{d}$ - can be described as
\begin{equation}
   \mathcal{L}_{g} = -\mathbb{E} [\mathcal{D}\left(\mathcal{G}\left(\textbf{I}\right)\right)] \quad and \quad
    \label{gan_loss_g}
    \vspace{-0.2cm}
\end{equation}
\begin{equation}
   \mathcal{L}_{d} = \mathbb{E}[\mathcal{D}\left(\mathcal{G}\left(\textbf{I}\right)\right)] - \mathbb{E}[\mathcal{D}\left(\textbf{I}^*\right)] + \lambda_{gp} \mathcal{E}_{gp}.
    \label{gan_loss_d}
\end{equation}

Unlike in a traditional GAN, WGAN's $\mathcal{G}$ and $\mathcal{D}$ are not trained for the same number of iterations. Multiple $\mathcal{D}$ iterations are performed for each $\mathcal{G}$ pass; we follow the official WGAN advice and set the $\mathcal{D}$:$\mathcal{G}$ iterations ratio to 5:1. Finally, while the pix2pixHD baseline uses InstanceNorm in both the generator and discriminator, our \ours\ instead uses learnable affine parameters with the normalisation layers in $\mathcal{D}$. 

\section{Experiments}
The proposed \ours\ was written in PyTorch. The loss scaling values mentioned in the Methodology were set as follows: $\lambda_{gp}$ = 10, $\lambda_{os}$ = 1, $\lambda_{perc}$ = 2, $\lambda_{sfr}$ = 5, $\lambda_{feat}$ = 2 and $\lambda_{id}$ = 1. The model was trained for 30 epochs and the best checkpoint from this range was chosen. Adam was used as the optimiser, with betas set to (0.0, 0.9). During training, we used the StepLR scheduler with an initial learning rate of $5 \times 10^{-4}$, a step of 10 and a gamma of 0.1.

For the ablation study (Sec.\ \ref{ablation_main}) and the first part of the experiments (Sec.\ \ref{istd_section}), all models were trained and evaluated on full-size images (640$\times$480) from the ISTD \cite{wang_stacked_2018} dataset. The training set contains 1331 samples, yet due to our two-branch approach, we form approx.\ 10.3k training pairs. The test set consists of 540 images which are fed into our generator individually, as the technique does not require multiple inputs at test time. We also train and test our model on the adjusted ISTD (AISTD) dataset \cite{le_2019_via} (Sec.\ \ref{aistd_section}) with the same parameters.

In the following sections the performance of \ours\ and other models is evaluated qualitatively and quantitatively. To follow the standard practise in the shadow removal domain, the numerical evaluation is presented in terms of RMSE calculated in the Lab colour space. However, we note that this is a mislabelling by previous authors, and that the evaluation script used by all cited authors actually calculates the error in terms of MAE (mean absolute error). This is a known, previously identified error in the literature, \eg\ \cite{mae_1,hu_mask-shadowgan_2019}. To follow the standards and facilitate future comparisons, we also report MAE values but label them as RMSE in the relevant tables.

\begin{table}[b]
\small
\vspace{-0.3cm}
\caption{Loss ablation study: impact of gradual loss addition on the performance of the proposed \ours. A/A = ``as above". }
\vspace{-0.3cm}
\begin{center}
\begin{tabular}{cccc}
\hline
%\multicolumn{3}{|c|}{s} \\
%\hline  \hline
Method & RMSE(A) & RMSE(S) & RMSE(N) \\
\hline
base & 13.17 & 15.84 & 12.82 \\
base + $\mathcal{L}_{sfr}$ & 7.80 & \underline{12.59} & 7.12 \\
base + $\mathcal{L}_{id}$ & 7.66 & 16.91 & \underline{6.15} \\
base + $\mathcal{L}_{sfr}$ + $\mathcal{L}_{id}$ & \textbf{7.09} & 14.99 & \textbf{5.94} \\
\hline
A/A + $\mathcal{L}_{perc}$  + $\mathcal{L}_{feat}$ & \underline{7.12} & \textbf{12.16} & 6.38 \\
\hline
\end{tabular}
\end{center} 
\label{loss_ablation}
\vspace{-0.5cm}
\end{table}

\subsection{Ablation study}
\label{ablation_main}
In this section we review the model's losses and demonstrate the compactness of the proposed \ours.

\subsubsection{Loss ablation study} 
\label{ablation}
We first evaluate the influence of each model loss on the overall system performance. During the ablation tests we do not remove the GAN losses - $\mathcal{L}_g$ and $\mathcal{L}_d$ - as well as the output similarity $\mathcal{L}_{os}$ as they are absolutely crucial to our model; we denote this case as `base'. This is then expanded by gradually adding the other model losses. The differences stemming from each model change are shown in Fig.\ \ref{loss_ablation_fig} and Table \ref{loss_ablation}. 

\begin{figure*}[h]
\begin{center}
\footnotesize
\begin{tabular}{ c @{\hspace{0.05cm}} c @{\hspace{0.05cm}} c @{\hspace{0.05cm}} c @{\hspace{0.05cm}} c @{\hspace{0.05cm}} c @{\hspace{0.05cm}} c }
 Input &Mask-ShadowGAN & LG-ShadowNet & DC-ShadowNet & \ours\ (ours) & Reference \\

\includegraphics[width=2.5cm]{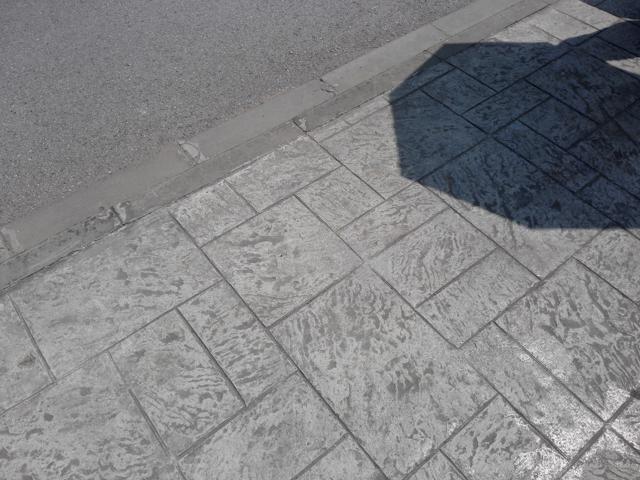}&
\includegraphics[width=2.5cm]{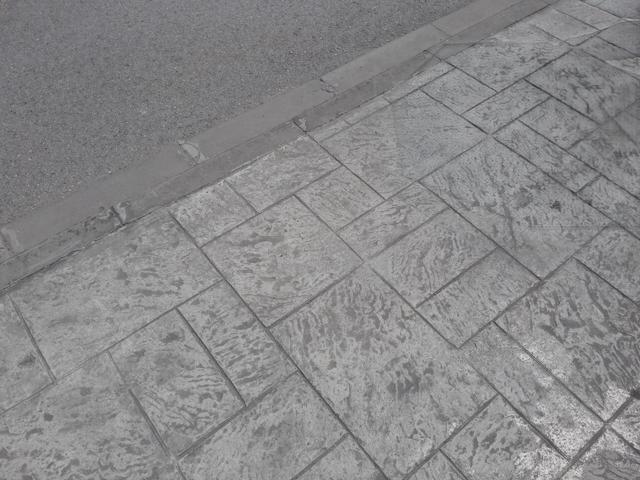}&
\includegraphics[width=2.5cm]{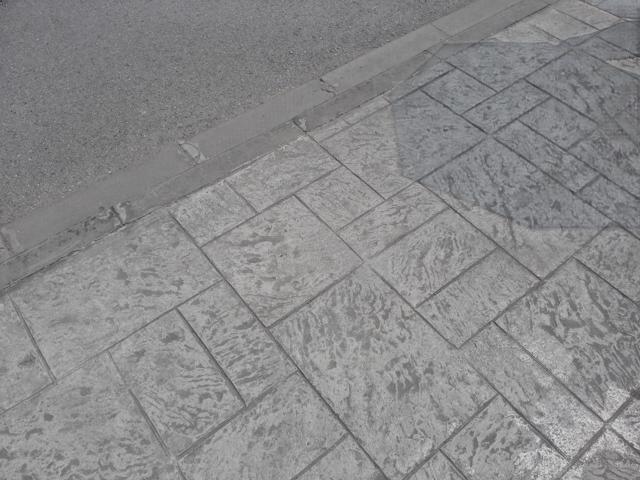}&
\includegraphics[width=2.5cm]{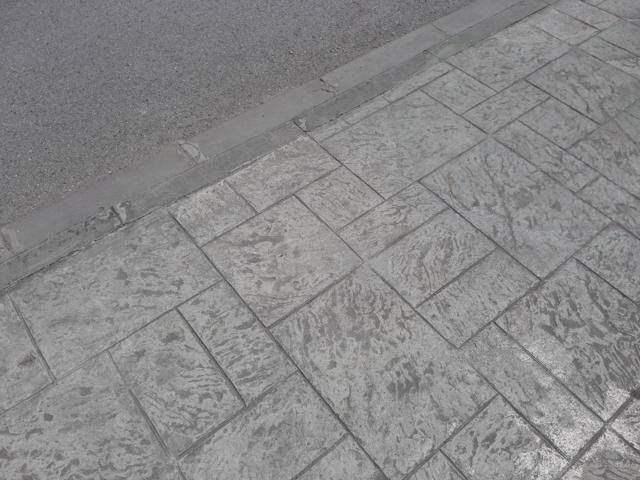}&
\includegraphics[width=2.5cm]{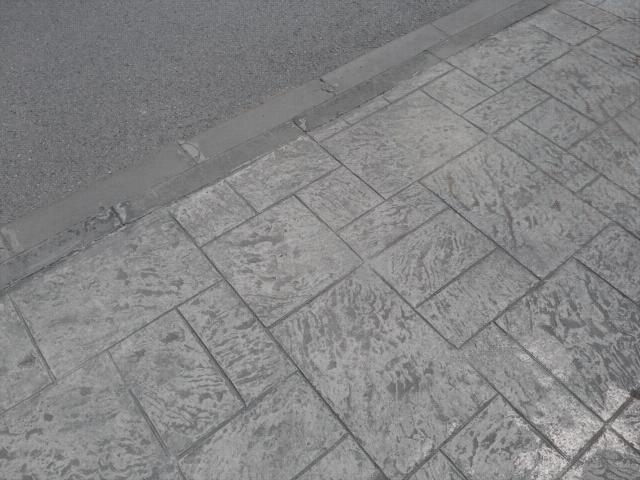}&
\includegraphics[width=2.5cm]{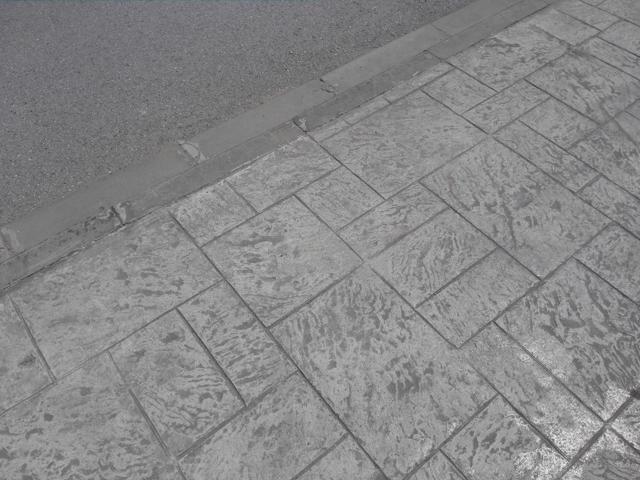}\\

\includegraphics[width=2.5cm]{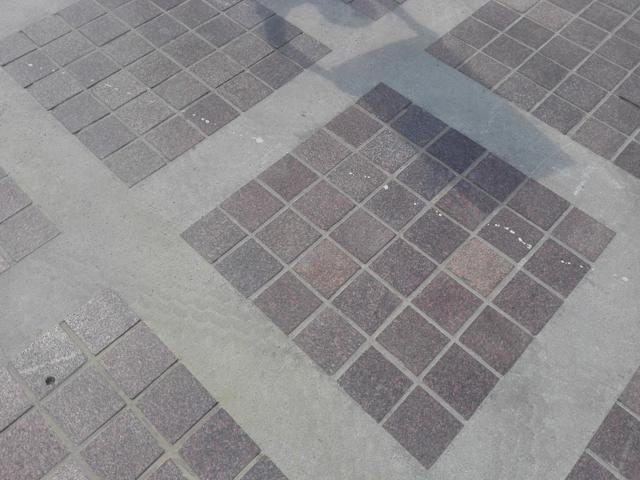}&
\includegraphics[width=2.5cm]{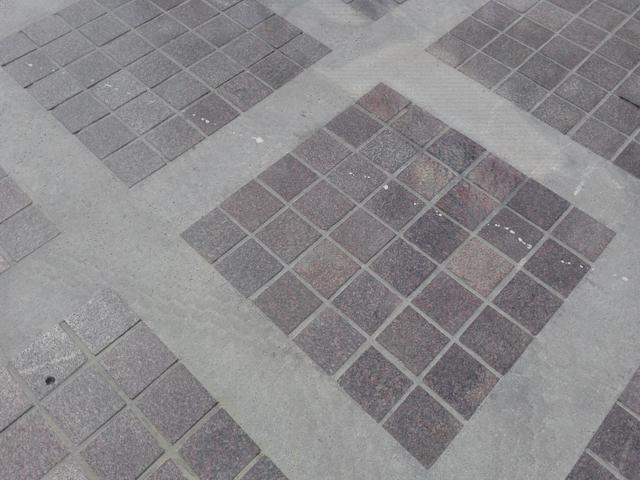}&
\includegraphics[width=2.5cm]{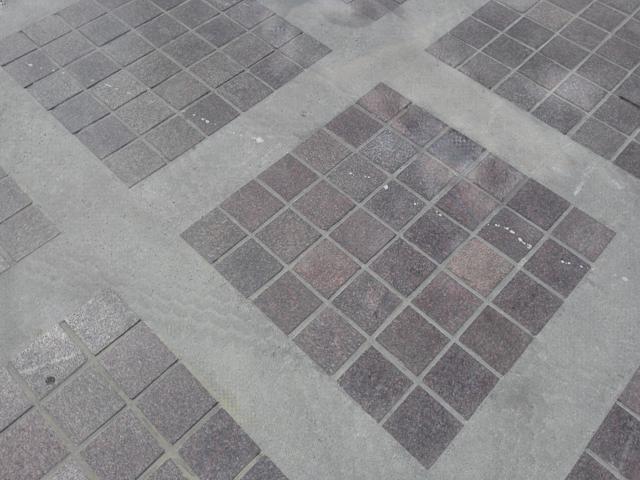}&
\includegraphics[width=2.5cm]{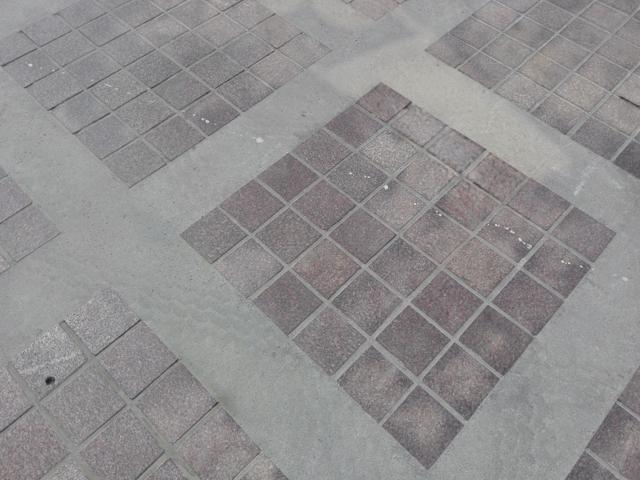}&
\includegraphics[width=2.5cm]{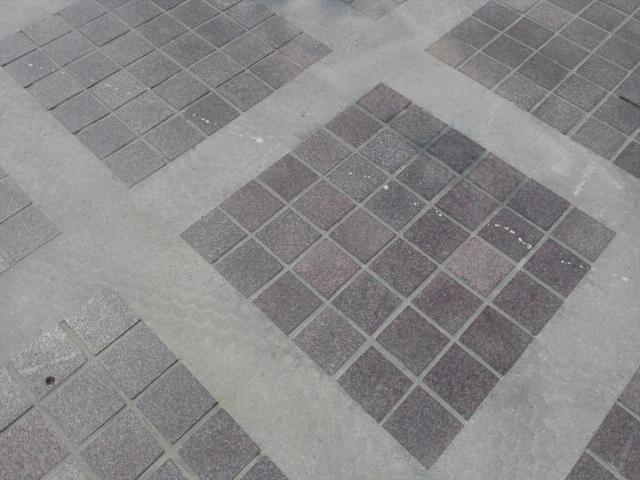}&
\includegraphics[width=2.5cm]{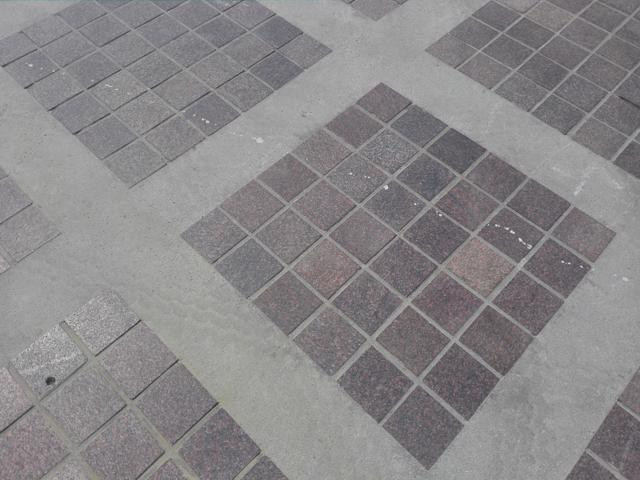}\\

\includegraphics[width=2.5cm]{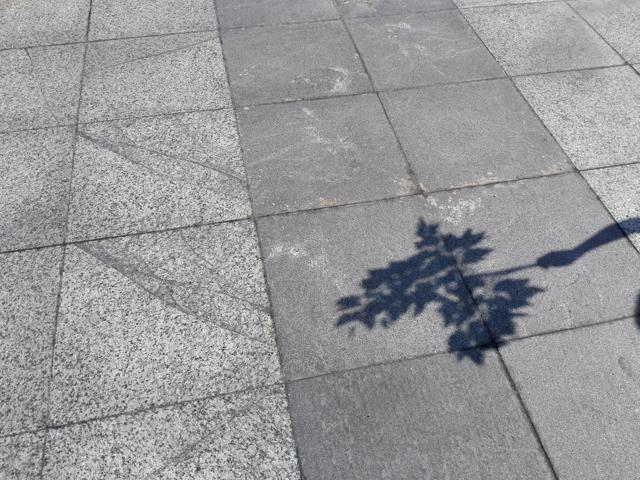}&
\includegraphics[width=2.5cm]{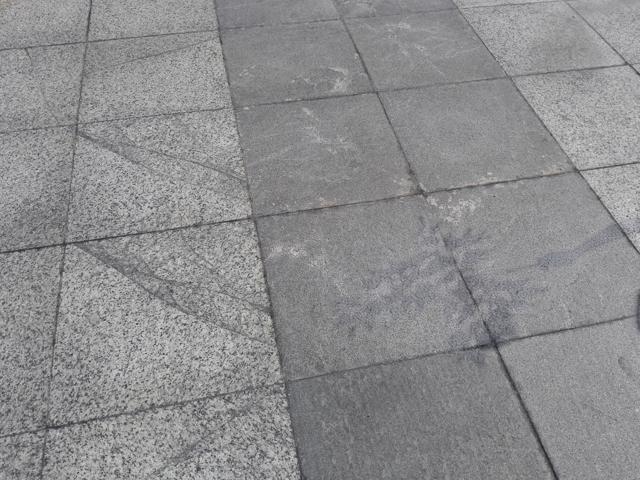}&
\includegraphics[width=2.5cm]{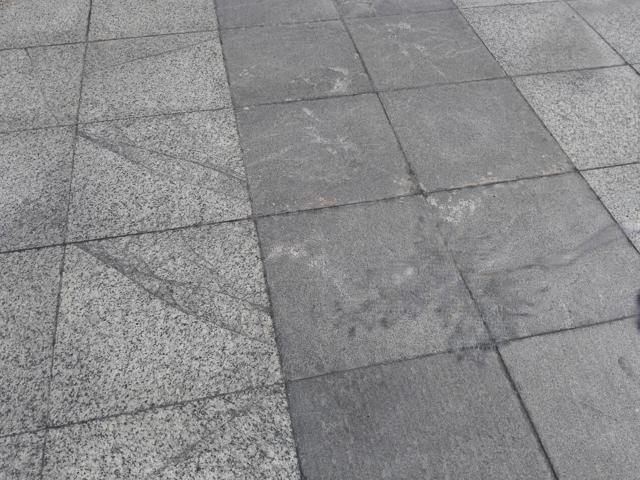}&
\includegraphics[width=2.5cm]{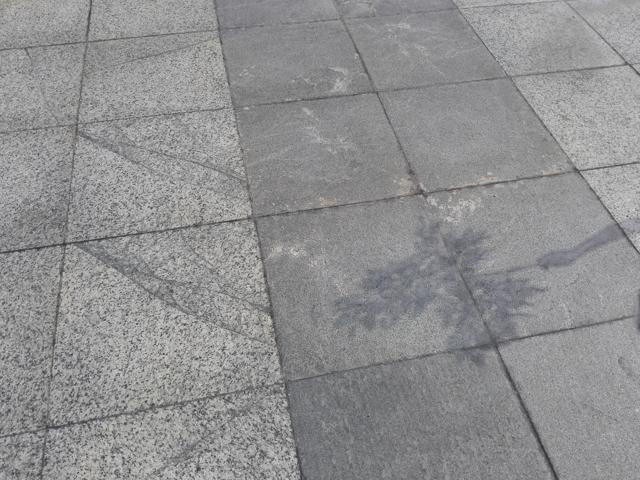}&
\includegraphics[width=2.5cm]{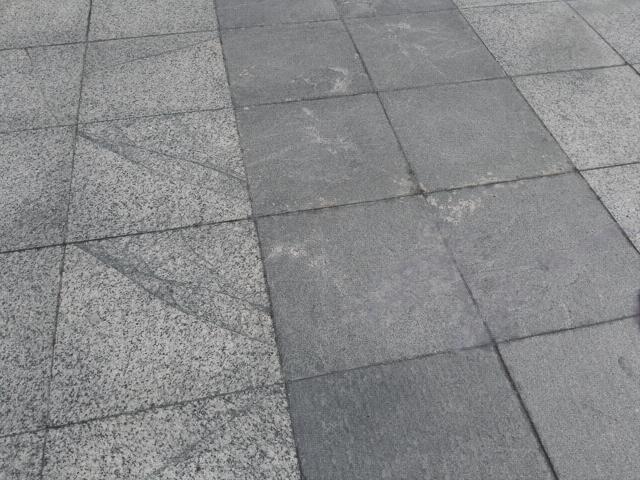}&
\includegraphics[width=2.5cm]{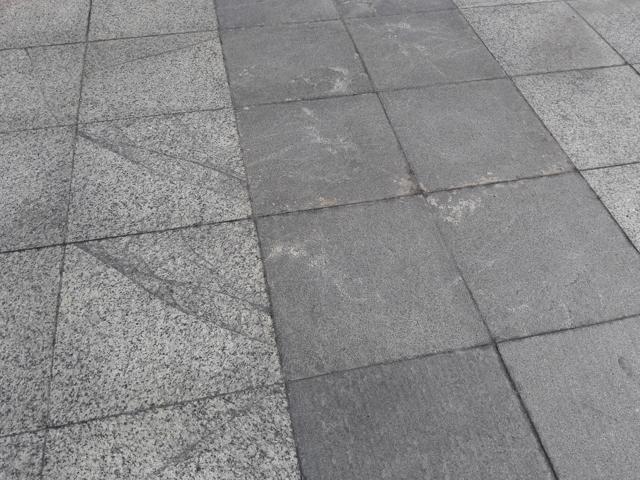}\\

\end{tabular}
\end{center}
\vspace{-0.2cm}
\caption{Visual results on the ISTD dataset.  }
\vspace{-0.2cm}
\label{istd_fig}
\end{figure*}

In the base case scenario, the error is spread between the shadow (S) and shadow-free (N) regions. The resulting images have some artefacts and their colours are slightly muted. The introduction of $\mathcal{L}_{sfr}$ leads to a significant drop in the shadow-free region error, which perfectly demonstrates the loss's purpose; the general quality also improves. $\mathcal{L}_{id}$ halves the shadow-free region error and leads to a slight increase in RMSE(S). The goal of identity loss is to prevent changes from being made to the shadow-free region. Since the loss is calculated on fully shadow-free samples, it does not introduce any awareness of shadows and, thus, might decrease performance in this area. Combining both of the newly introduced losses further improves the the RMSE(N) and results in an RMSE(S) score somewhere between the others'. Finally, we add the feature-based losses  $\mathcal{L}_{perc}$ and $\mathcal{L}_{feat}$. The losses do not have a significant impact on the numerical results yet they improve the visual quality of the outputs and, in particular, decrease the appearance of shadow edges due to their robustness to misalignment (zoom in on bottom row). This may contribute to the slight drop in RMSE(S) in the final version of the model.

\begin{figure}[b]
    \vspace{-0.5cm}
    \includegraphics[width=8.2cm]{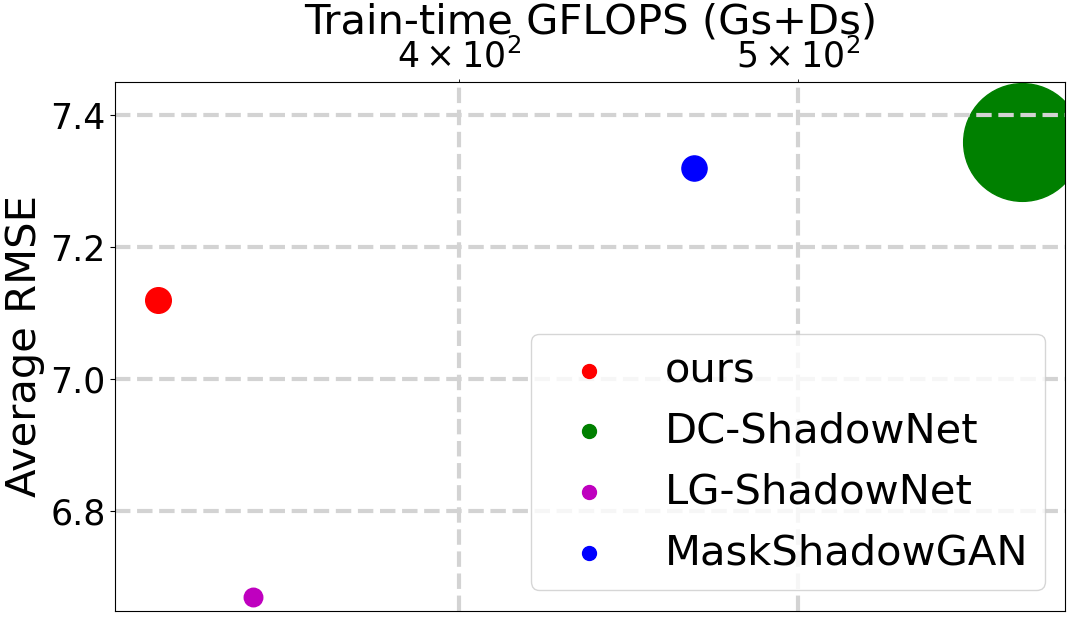}
    \caption{Model error vs train-time GFLOPS comparison between each model's generator(s)+discriminator(s) trained on full-size ISTD images. Circle radius is proportional to the total number of model parameters.}
    \label{flops}
\end{figure}

\subsubsection{Model compactness study}
\label{compactness_section}
Recently, the growing accessibility of powerful GPUs has accelerated the development of top-accuracy models. However, this improvement is usually coupled with an increase in model size and computational requirements. Therefore, in this section, we wish to show that the proposed \ours\ achieves good performance without excessively inflating the network. 

The results of our study are visualised in Fig.\ \ref{flops}. The graph plots the models' performance in terms of RMSE(A) vs a total number of train-time GFLOPS. The marker used to represent each solution has a radius corresponding to its total number of parameters. The presented values represent the GFLOPS/parameters of the generator and discriminator networks forming a given de-shadowing system.

In this comparison we consider 3 self-supervised, cycle-consistency based models: Mask-ShadowGAN \cite{hu_mask-shadowgan_2019}, LG-ShadowNet \cite{liu_shadow_2021} and DC-ShadowNet \cite{jin_2021_unsup}. We take the numerical RMSE results for the first two models from their papers and run the outputs of the pre-trained DC-ShadowNet through the official evaluation script (the authors only report results on cropped data). Our \ours\ is the least computationally expensive self-supervised shadow removal model reviewed here. The network also comes second in terms of RMSE(A), just after LG-ShadowNet. In terms of model parameters, the SqueezeNet-based LG-ShadowNet is closely followed by our \ours\ and Mask-ShadowGAN. The most recent DC-ShadowNet is the largest and most computationally-heavy model in our evaluation.

\begin{figure*}[t]
\begin{center}
\footnotesize
\begin{tabular}{ c @{\hspace{0.05cm}} c @{\hspace{0.05cm}} c @{\hspace{0.05cm}} c @{\hspace{0.05cm}} c @{\hspace{0.05cm}} c @{\hspace{0.05cm}} c }
 Input &Mask-ShadowGAN & LG-ShadowNet & DC-ShadowNet &\ours\ (ours) & Reference \\
\includegraphics[width=2.5cm]{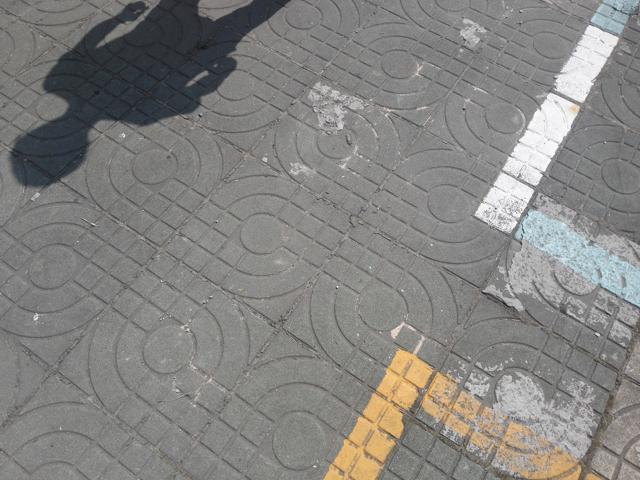}&
\includegraphics[width=2.5cm]{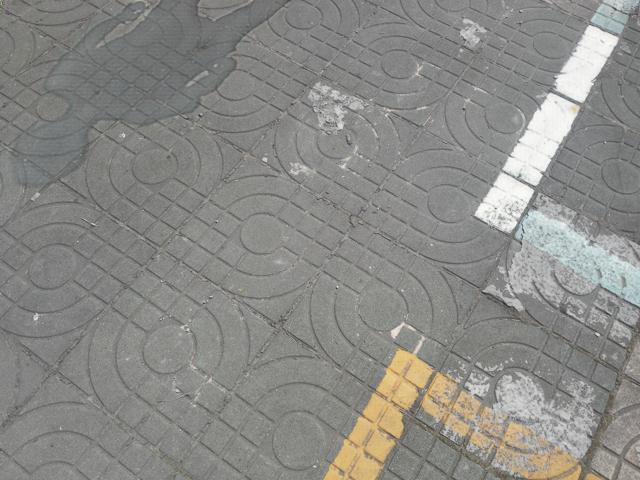}&
\includegraphics[width=2.5cm]{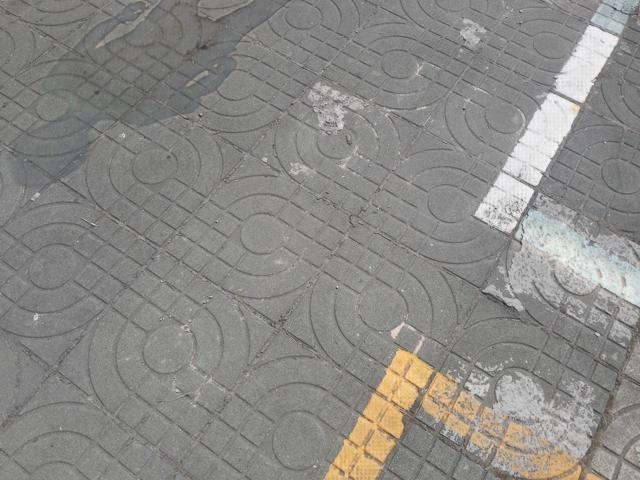}&
\includegraphics[width=2.5cm]{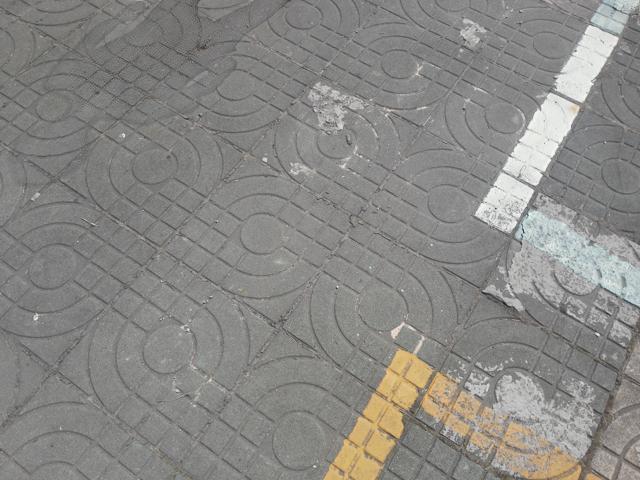}&
\includegraphics[width=2.5cm]{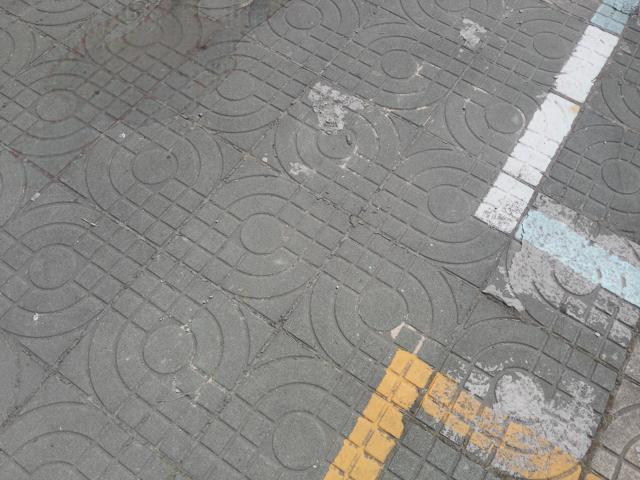}&
\includegraphics[width=2.5cm]{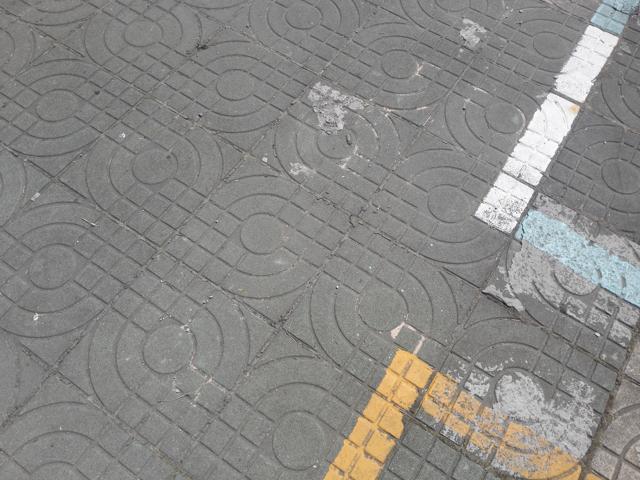}\\

\includegraphics[width=2.5cm]{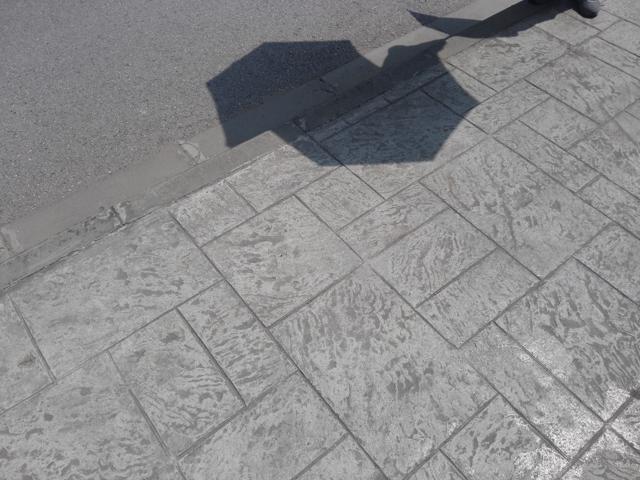}&
\includegraphics[width=2.5cm]{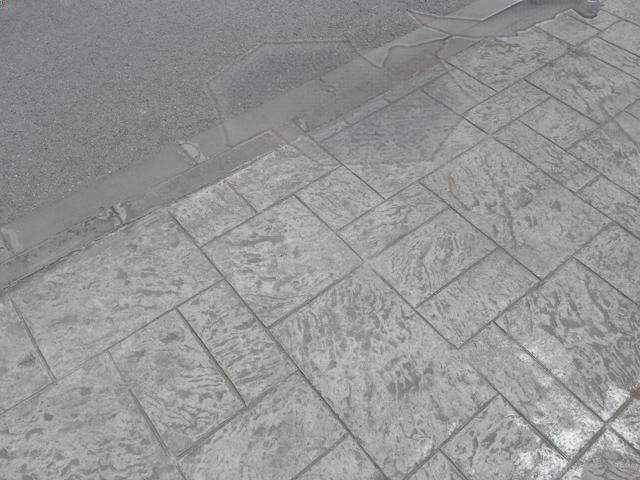}&
\includegraphics[width=2.5cm]{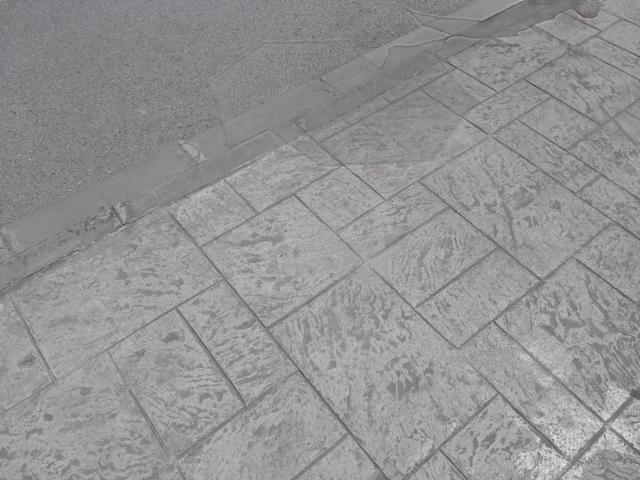}&
\includegraphics[width=2.5cm]{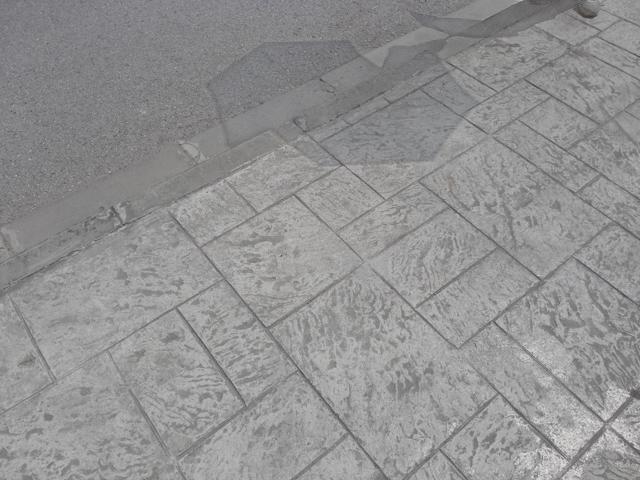}&
\includegraphics[width=2.5cm]{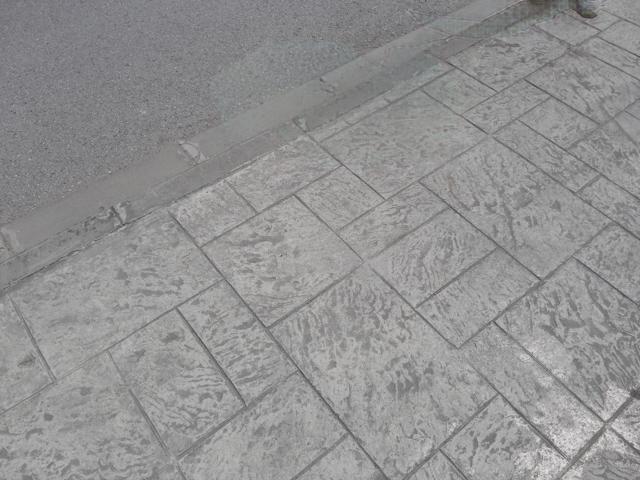}&
\includegraphics[width=2.5cm]{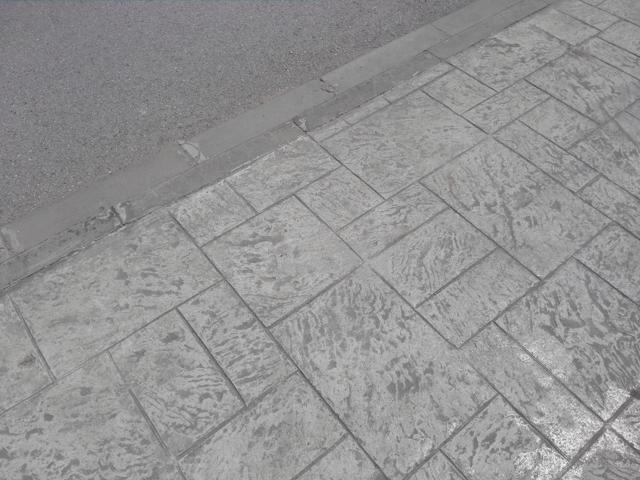}\\

\includegraphics[width=2.5cm]{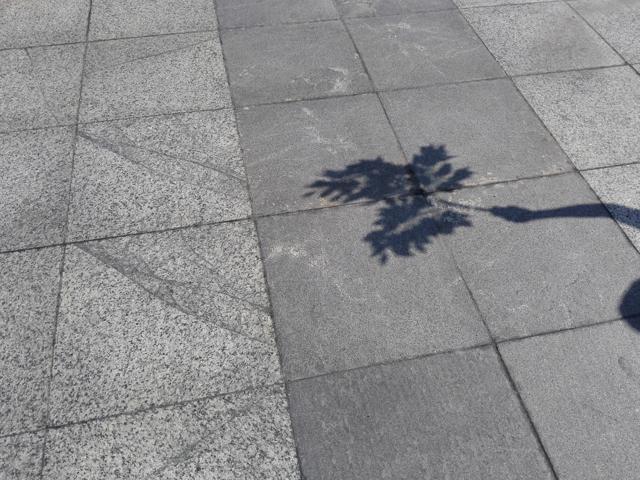}&
\includegraphics[width=2.5cm]{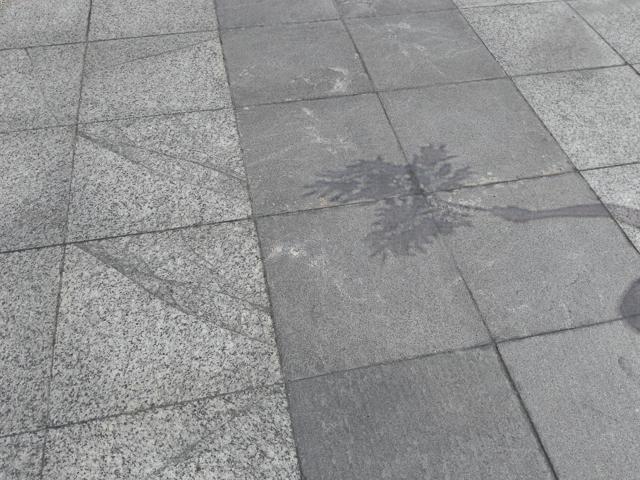}&
\includegraphics[width=2.5cm]{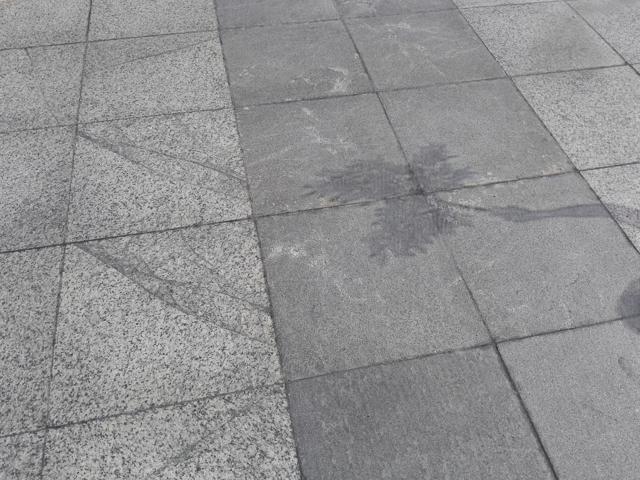}&
\includegraphics[width=2.5cm]{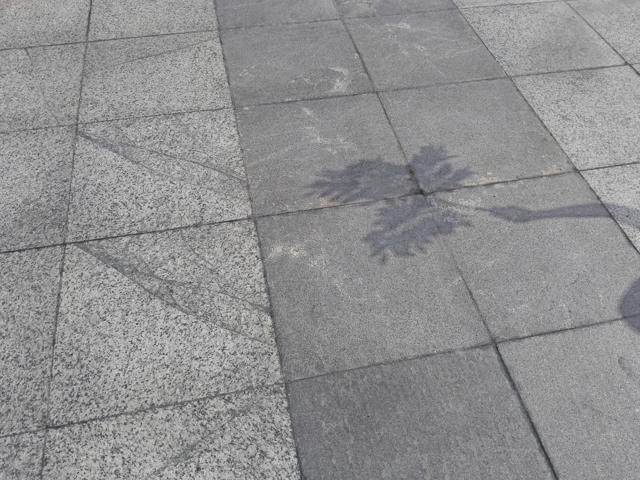}&
\includegraphics[width=2.5cm]{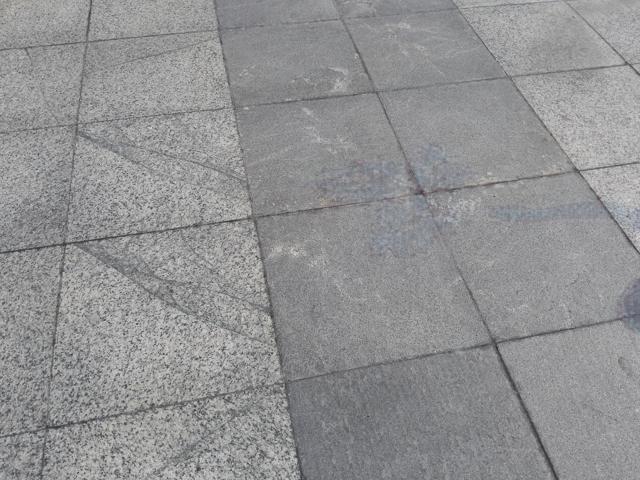}&
\includegraphics[width=2.5cm]{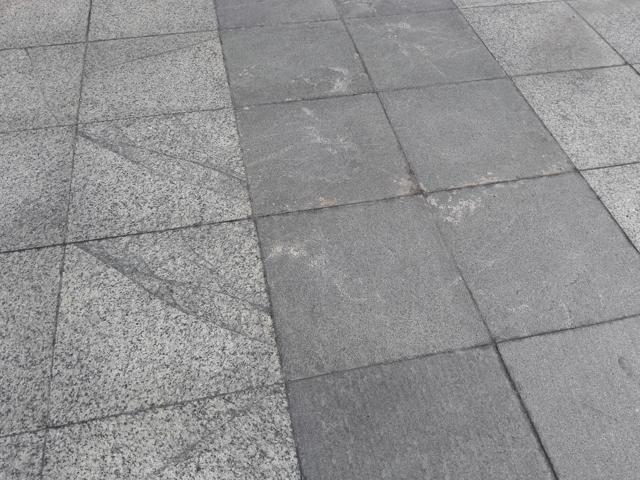}\\

\end{tabular}
\end{center}
\vspace{-0.5cm}
\caption{Visual results on the AISTD dataset.  }
\label{aistd_fig}
\end{figure*}

\subsection{Results on ISTD}
\label{istd_section}
Next, we undertake a more in-depth comparison of our \ours\ model with the relevant self-supervised state-of-the-art methods introduced in Sec.\ \ref{compactness_section}. The images used for the comparisons were provided by the authors (LG-ShadowNet) or generated using the published pre-trained models (the other 2 methods).

The results of this state-of-the-art experiment are presented in Table \ref{istd} and Fig.\ \ref{istd_fig}. Our model is outperformed only by LG-ShadowNet, which surpasses all other models numerically. However, when it comes to visual results, both it and Mask-ShadowGAN struggle with shadow edges. These are less prominent in outputs generated with DC-ShadowNet and even less so in our \ours. We attribute this to the use of perceptual losses, taking care of the overall scene structure alongside the removal. 

Additionally, we note that the competitors' models have a tendency to unnecessarily lighten darker image regions. This is particularly clear in the middle row of Fig.\ \ref{istd_fig}, where a number of floor tiles have their centres lightened (DC-ShadowNet) or small cloud-like areas appear (LG-ShadowNet). The issue does not seem to be present in \ours\ outputs.

\begin{table}[t]
\setlength{\tabcolsep}{4pt}
\small
\vspace{-0.3cm}
\caption{Results on the ISTD dataset (full-size images). }
\vspace{-0.4cm}
\begin{center}
\begin{tabular}{cccc}
\hline
Method & RMSE(A) & RMSE(S) & RMSE(N) \\
\hline 
Mask-ShadowGAN \cite{hu_mask-shadowgan_2019} & 7.32& 12.65 & 6.57 \\
LG-ShadowNet \cite{liu_shadow_2021}& \textbf{6.67}& \underline{11.63}& \textbf{5.91}\\
DC-ShadowNet \cite{jin_2021_unsup} & 7.36 &	\textbf{11.21} & 6.64\\
\ours\ (ours) &\underline{7.12} & 12.16 & \underline{6.38}  \\
\hline
\end{tabular}
\end{center} 
\label{istd}
\vspace{-0.6cm}
\end{table}

\begin{table}[b]
\setlength{\tabcolsep}{4pt}
\small
\vspace{-0.3cm}
\caption{Results on the AISTD dataset (full-size images). }
\vspace{-0.4cm}
\begin{center}

\begin{tabular}{cccc}
\hline
Method & RMSE(A) & RMSE(S) & RMSE(N) \\
\hline 
Mask-ShadowGAN \cite{hu_mask-shadowgan_2019} & 5.84& \underline{12.28} & 4.82 \\
LG-ShadowNet \cite{liu_shadow_2021}& \textbf{5.02}& \textbf{10.64}& \textbf{4.02}\\
DC-ShadowNet \cite{jin_2021_unsup} & \underline{5.64}&	12.63 &\underline{4.33} \\
\ours\ (ours) & 5.71 & 12.86 & 4.43 \\
\hline
\end{tabular}
\end{center} 
\label{aistd}
\vspace{-0.7cm}
\end{table}

\subsection{Results on AISTD}
\label{aistd_section}
We retrain our model and test it on the AISTD \cite{le_2019_via} dataset. Once again, we compare \ours\ with SOTA on full-size images, \ie\ 640$\times$480 pixels. Mask-ShadowGAN was not trained on AISTD, so we retrain the model using the code from the official github repo. The numerical and visual results for LG-Shadow come directly from the authors' github and paper. Finally, DC-ShadowNet reports performance only on cropped data, so again we use their pre-trained model and the original evaluation script to get the numerical results for full-size inputs. 

The results of this comparison are presented in Table \ref{aistd} and Fig.\ \ref{aistd_fig}. Just like before, LG-ShadowNet is the top performer in the AISTD comparison, with the remaining 3 models achieving similar performance. In Fig.\ \ref{aistd_fig} we can see that the models, including the top-performing LG-ShadowNet, suffer from similar faults as on the ISTD dataset: In the top 2 rows, the outputs generated by our competitors have clear shadow edges while our reconstructions have significantly smoother shadow boundaries. The appearance of the shadow fill is also most visibly reduced in the \ours\ samples. Finally, our model can successfully detect and remove shadow in cases where other models fail (bottom row). Despite coming third quantitatively, we believe we have shown that \ours\ can generate results that look the most pleasant to the human visual system.

\section{Conclusions \& future work}
In this paper we presented \ours\ -- a shadow removal model with a novel \textit{\approach} approach to self-supervision. The proposed network achieves similar quantitative performance to the state-of-the-art self-supervised frameworks at a low computational cost. Additionally, and most importantly, we demonstrate superior qualitative performance: \ours\ generates de-shadowed images with virtually imperceptible shadows, both in terms of edges as well as the inside fill, while maintaining the scene colours.

In future work, it would be interesting to further research the problem of colour adjustments in the shadow region. Adding the feature-based losses, we managed to address a common issue of leftover shadow edges. However, matching the colour of the de-shadowed region to the rest of the scene is still an ongoing problem -- both for us and the other models. Moreover, the proposed \ours\ does not need ground truth for training, yet the method still relies on paired input data. Therefore, we could investigate ways of further reducing our supervision requirements, \eg\ by creatively using data augmentations. Finally, the (A)ISTD dataset exhibits some misalignment between shadowed and shadow-free samples. Our \ours\ deals with this well due a mix of pixel-wise as well as perceptual losses. However, during model development we have not experimented with any higher levels of misalignment, so it is unclear how the network would perform given worse quality data.
%\vspace{0.3cm}
\\[5pt]
\fbox{\begin{minipage}{23em}
\footnotesize This work was partially supported by the BBC and the EPSRC's iCASE project ``Computational lighting in video" (voucher 19000034).
\end{minipage}}

\clearpage

{
    \small
    \bibliographystyle{ieeenat_fullname}
    \bibliography{main}

\begin{thebibliography}{61}
\providecommand{\natexlab}[1]{#1}
\providecommand{\url}[1]{\texttt{#1}}
\expandafter\ifx\csname urlstyle\endcsname\relax
  \providecommand{\doi}[1]{doi: #1}\else
  \providecommand{\doi}{doi: \begingroup \urlstyle{rm}\Url}\fi

\bibitem[Arjovsky et~al.(2017)Arjovsky, Chintala, and Bottou]{arjovsky_2017_wgan}
Martin Arjovsky, Soumith Chintala, and L{\'e}on Bottou.
\newblock Wasserstein generative adversarial networks.
\newblock In \emph{International Conference on Machine Learning (ICML)}, 2017.

\bibitem[Barron and Malik(2014)]{barron_2014_shape}
Jonathan~T Barron and Jitendra Malik.
\newblock Shape, illumination, and reflectance from shading.
\newblock \emph{IEEE transactions on pattern analysis and machine intelligence}, 37\penalty0 (8), 2014.

\bibitem[Chen et~al.(2021)Chen, Long, Zhang, and Xiao]{chen_2021_canet}
Zipei Chen, Chengjiang Long, Ling Zhang, and Chunxia Xiao.
\newblock Canet: A context-aware network for shadow removal.
\newblock In \emph{Proceedings of the IEEE/CVF International Conference on Computer Vision (ICCV)}, 2021.

\bibitem[Cun et~al.(2020)Cun, Pun, and Shi]{cun_2020_HAN}
Xiaodong Cun, Chi-Man Pun, and Cheng Shi.
\newblock Towards ghost-free shadow removal via dual hierarchical aggregation network and shadow matting gan.
\newblock In \emph{Proceedings of the AAAI Conference on Artificial Intelligence (AAAI)}, 2020.

\bibitem[Ding et~al.(2019)Ding, Long, Zhang, and Xiao]{ding_argan_2019}
Bin Ding, Chengjiang Long, Ling Zhang, and Chunxia Xiao.
\newblock Argan: Attentive recurrent generative adversarial network for shadow detection and removal.
\newblock In \emph{Proceedings of the IEEE/CVF International Conference on Computer Vision (ICCV)}, 2019.

\bibitem[Drew et~al.(2003)Drew, Finlayson, and Hordley]{drew_2003_recovery}
Mark~S Drew, Graham~D Finlayson, and Steven~D Hordley.
\newblock Recovery of chromaticity image free from shadows via illumination invariance.
\newblock In \emph{IEEE Workshop on Color and Photometric Methods in Computer Vision (ICCVW)}, 2003.

\bibitem[Eigen et~al.(2014)Eigen, Puhrsch, and Fergus]{eigen_2014_depth}
David Eigen, Christian Puhrsch, and Rob Fergus.
\newblock Depth map prediction from a single image using a multi-scale deep network.
\newblock \emph{Advances in neural information processing systems}, 27, 2014.

\bibitem[Finlayson and Drew(2001)]{finlayson_2001_cameracalib}
G.D. Finlayson and M.S. Drew.
\newblock 4-sensor camera calibration for image representation invariant to shading, shadows, lighting, and specularities.
\newblock In \emph{Proceedings of the IEEE International Conference on Computer Vision (ICCV)}, 2001.

\bibitem[Fu et~al.()Fu, Zhou, Guo, Juefei-Xu, Yu, Feng, Liu, and Wang]{mae_1}
Lan Fu, Changqing Zhou, Qing Guo, Felix Juefei-Xu, Hongkai Yu, Wei Feng, Yang Liu, and Song Wang.
\newblock {Auto-exposure fusion for single-image shadow removal} - official github repo.
\newblock https://github.com/tsingqguo/exposure-fusion-shadow-removal.
\newblock Accessed: 2022-03-03.

\bibitem[Fu et~al.(2021)Fu, Zhou, Guo, Juefei-Xu, Yu, Feng, Liu, and Wang]{fu_2021_auto}
Lan Fu, Changqing Zhou, Qing Guo, Felix Juefei-Xu, Hongkai Yu, Wei Feng, Yang Liu, and Song Wang.
\newblock Auto-exposure fusion for single-image shadow removal.
\newblock In \emph{Proceedings of the {IEEE}/{CVF} {Conference} on {Computer} {Vision} and {Pattern} {Recognition} ({CVPR})}, 2021.

\bibitem[Georgiadis et~al.(2023)Georgiadis, Yucel, Skartados, Dimaridou, Drosou, Saa-Garriga, and Manganelli]{georgiadis2023lp}
Konstantinos Georgiadis, M~Kerim Yucel, Evangelos Skartados, Valia Dimaridou, Anastasios Drosou, Albert Saa-Garriga, and Bruno Manganelli.
\newblock Lp-ioanet: Efficient high resolution document shadow removal.
\newblock In \emph{ICASSP 2023-2023 IEEE International Conference on Acoustics, Speech and Signal Processing (ICASSP)}, 2023.

\bibitem[Gong and Cosker(2016)]{gong_2016_interactive}
Han Gong and Darren Cosker.
\newblock Interactive removal and ground truth for difficult shadow scenes.
\newblock \emph{JOSA A}, 33\penalty0 (9), 2016.

\bibitem[Goodfellow et~al.(2014)Goodfellow, Pouget-Abadie, Mirza, Xu, Warde-Farley, Ozair, Courville, and Bengio]{goodfellow_2014_gan}
Ian Goodfellow, Jean Pouget-Abadie, Mehdi Mirza, Bing Xu, David Warde-Farley, Sherjil Ozair, Aaron Courville, and Yoshua Bengio.
\newblock Generative adversarial nets.
\newblock \emph{Advances in neural information processing systems (NIPS)}, 27, 2014.

\bibitem[Gryka et~al.(2015)Gryka, Terry, and Brostow]{gryka_2015_softshadows}
Maciej Gryka, Michael Terry, and Gabriel~J. Brostow.
\newblock Learning to remove soft shadows.
\newblock \emph{ACM Transactions on Graphics}, 34\penalty0 (5), 2015.

\bibitem[Gulrajani et~al.(2017)Gulrajani, Ahmed, Arjovsky, Dumoulin, and Courville]{gulrajani_2017_wgan_gp}
Ishaan Gulrajani, Faruk Ahmed, Martin Arjovsky, Vincent Dumoulin, and Aaron~C Courville.
\newblock Improved training of wasserstein gans.
\newblock \emph{Advances in neural information processing systems}, 30, 2017.

\bibitem[Guo et~al.(2023{\natexlab{a}})Guo, Huang, Liu, Cheng, and Wen]{guo2023shadowformer}
Lanqing Guo, Siyu Huang, Ding Liu, Hao Cheng, and Bihan Wen.
\newblock Shadowformer: Global context helps image shadow removal.
\newblock \emph{arXiv preprint arXiv:2302.01650}, 2023{\natexlab{a}}.

\bibitem[Guo et~al.(2023{\natexlab{b}})Guo, Wang, Yang, Huang, Wang, Pfister, and Wen]{guo2023shadowdiffusion}
Lanqing Guo, Chong Wang, Wenhan Yang, Siyu Huang, Yufei Wang, Hanspeter Pfister, and Bihan Wen.
\newblock Shadowdiffusion: When degradation prior meets diffusion model for shadow removal.
\newblock In \emph{Proceedings of the IEEE/CVF Conference on Computer Vision and Pattern Recognition (CVPR)}, 2023{\natexlab{b}}.

\bibitem[Guo et~al.(2023{\natexlab{c}})Guo, Wang, Yang, Wang, and Wen]{guo2023boundary}
Lanqing Guo, Chong Wang, Wenhan Yang, Yufei Wang, and Bihan Wen.
\newblock Boundary-aware divide and conquer: A diffusion-based solution for unsupervised shadow removal.
\newblock In \emph{Proceedings of the IEEE/CVF International Conference on Computer Vision (ICCV)}, 2023{\natexlab{c}}.

\bibitem[He et~al.(2021)He, Xing, Zhang, and Chen]{he_2021_unsupervised}
Yingqing He, Yazhou Xing, Tianjia Zhang, and Qifeng Chen.
\newblock Unsupervised portrait shadow removal via generative priors.
\newblock In \emph{ACM International Conference on Multimedia (ACM MM)}, 2021.

\bibitem[Howard(2012)]{howard_2012_depth}
Ian~P Howard.
\newblock \emph{Perceiving in depth, volume 1: basic mechanisms}.
\newblock Oxford University Press, 2012.

\bibitem[Hu et~al.(2019)Hu, Jiang, Fu, and Heng]{hu_mask-shadowgan_2019}
Xiaowei Hu, Yitong Jiang, Chi-Wing Fu, and Pheng-Ann Heng.
\newblock Mask-{ShadowGAN}: {Learning} to {Remove} {Shadows} from {Unpaired} {Data}.
\newblock In \emph{Proceedings of the IEEE/CVF International Conference on Computer Vision (ICCV)}, 2019.

\bibitem[Hu et~al.(2020)Hu, Fu, Zhu, Qin, and Heng]{hu_2020_context}
Xiaowei Hu, Chi-Wing Fu, Lei Zhu, Jing Qin, and Pheng-Ann Heng.
\newblock Direction-aware spatial context features for shadow detection and removal.
\newblock \emph{IEEE Transactions on Pattern Analysis and Machine Intelligence}, 42\penalty0 (11), 2020.

\bibitem[Jin et~al.(2021)Jin, Sharma, and Tan]{jin_2021_unsup}
Yeying Jin, Aashish Sharma, and Robby~T Tan.
\newblock Dc-shadownet: Single-image hard and soft shadow removal using unsupervised domain-classifier guided network.
\newblock In \emph{Proceedings of the IEEE/CVF International Conference on Computer Vision (ICCV)}, 2021.

\bibitem[Jin et~al.(2022)Jin, Yang, Ye, Yuan, and Tan]{jin2022des3}
Yeying Jin, Wenhan Yang, Wei Ye, Yuan Yuan, and Robby~T Tan.
\newblock Des3: Attention-driven self and soft shadow removal using vit similarity and color convergence.
\newblock \emph{arXiv preprint arXiv:2211.08089}, 2022.

\bibitem[Jung et~al.(2018)Jung, Hasan, and Kim]{jung_2018_documents}
Seungjun Jung, Muhammad~Abul Hasan, and Changick Kim.
\newblock Water-filling: An efficient algorithm for digitized document shadow removal.
\newblock In \emph{Proceedings of the Asian Conference on Computer Vision (ACCV)}, 2018.

\bibitem[Karras et~al.(2020)Karras, Laine, Aittala, Hellsten, Lehtinen, and Aila]{stylegan2}
Tero Karras, Samuli Laine, Miika Aittala, Janne Hellsten, Jaakko Lehtinen, and Timo Aila.
\newblock Analyzing and improving the image quality of {StyleGAN}.
\newblock In \emph{Proceedings of the IEEE/CVF Conference on Computer Vision and Pattern Recognition (CVPR)}, 2020.

\bibitem[Kubiak et~al.(2021)Kubiak, Mustafa, Phillipson, Jolly, and Hadfield]{kubiak_2021_silt}
Nikolina Kubiak, Armin Mustafa, Graeme Phillipson, Stephen Jolly, and Simon Hadfield.
\newblock Silt: Self-supervised lighting transfer using implicit image decomposition.
\newblock In \emph{Proceedings of the British Machine Vision Conference (BMVC)}, 2021.

\bibitem[Le and Samaras(2019)]{le_2019_via}
Hieu Le and Dimitris Samaras.
\newblock Shadow removal via shadow image decomposition.
\newblock In \emph{Proceedings of the IEEE/CVF IEEE/CVF International Conference on Computer Vision (ICCV)}, 2019.

\bibitem[Le and Samaras(2020)]{le_2020_shadow}
Hieu Le and Dimitris Samaras.
\newblock From shadow segmentation to shadow removal.
\newblock In \emph{Proceedings of the European Conference on Computer Vision (ECCV)}, 2020.

\bibitem[Li et~al.(2023{\natexlab{a}})Li, Guo, Abdelfattah, Lin, Feng, Tsang, and Wang]{li2023leveraging}
Xiaoguang Li, Qing Guo, Rabab Abdelfattah, Di Lin, Wei Feng, Ivor Tsang, and Song Wang.
\newblock Leveraging inpainting for single-image shadow removal.
\newblock In \emph{Proceedings of the IEEE/CVF International Conference on Computer Vision (ICCV)}, 2023{\natexlab{a}}.

\bibitem[Li et~al.(2023{\natexlab{b}})Li, Guo, Cai, Feng, Tsang, and Wang]{li2023learning}
Xiaoguang Li, Qing Guo, Pingping Cai, Wei Feng, Ivor Tsang, and Song Wang.
\newblock Learning restoration is not enough: Transfering identical mapping for single-image shadow removal.
\newblock \emph{arXiv preprint arXiv:2305.10640}, 2023{\natexlab{b}}.

\bibitem[Liu et~al.(2024)Liu, Ke, Xu, Liu, Wang, and Lau]{liu2024recasting}
Yuhao Liu, Zhanghan Ke, Ke Xu, Fang Liu, Zhenwei Wang, and Rynson Lau.
\newblock Recasting regional lighting for shadow removal.
\newblock In \emph{Proceedings of the AAAI Conference on Artificial Inteligence (AAAI)}, 2024.

\bibitem[Liu et~al.(2021{\natexlab{a}})Liu, Yin, Mi, Pu, and Wang]{liu_shadow_2021}
Zhihao Liu, Hui Yin, Yang Mi, Mengyang Pu, and Song Wang.
\newblock Shadow {Removal} by a {Lightness}-{Guided} {Network} with {Training} on {Unpaired} {Data}.
\newblock \emph{IEEE Trans. on Image Process.}, 30, 2021{\natexlab{a}}.

\bibitem[Liu et~al.(2021{\natexlab{b}})Liu, Yin, Wu, Wu, Mi, and Wang]{liu_2021_shadow_gen}
Zhihao Liu, Hui Yin, Xinyi Wu, Zhenyao Wu, Yang Mi, and Song Wang.
\newblock From shadow generation to shadow removal.
\newblock In \emph{Proceedings of the {IEEE}/{CVF} {Conference} on {Computer} {Vision} and {Pattern} {Recognition} ({CVPR})}, 2021{\natexlab{b}}.

\bibitem[Lu et~al.(2021)Lu, Cole, Dekel, Zisserman, Freeman, and Rubinstein]{lu_2021_omnimatte}
Erika Lu, Forrester Cole, Tali Dekel, Andrew Zisserman, William~T Freeman, and Michael Rubinstein.
\newblock Omnimatte: associating objects and their effects in video.
\newblock In \emph{Proceedings of the IEEE/CVF Conference on Computer Vision and Pattern Recognition (CVPR)}, 2021.

\bibitem[Mei et~al.(2024)Mei, Figueroa, Lin, Ding, Cohen, and Patel]{mei2024latent}
Kangfu Mei, Luis Figueroa, Zhe Lin, Zhihong Ding, Scott Cohen, and Vishal~M Patel.
\newblock Latent feature-guided diffusion models for shadow removal.
\newblock In \emph{Proceedings of the IEEE/CVF Winter Conference on Applications of Computer Vision (WACV)}, 2024.

\bibitem[Otsu(1979)]{otsu_1979_threshold}
Nobuyuki Otsu.
\newblock A threshold selection method from gray-level histograms.
\newblock \emph{IEEE transactions on systems, man, and cybernetics}, 9\penalty0 (1), 1979.

\bibitem[Qu et~al.(2017)Qu, Tian, He, Tang, and Lau]{qu_deshadownet_2017}
L. Qu, J. Tian, S. He, Y. Tang, and R.~W.~H. Lau.
\newblock {DeshadowNet}: {A} {Multi}-context {Embedding} {Deep} {Network} for {Shadow} {Removal}.
\newblock In \emph{Proceedings of the {IEEE/CVF} {Conference} on {Computer} {Vision} and {Pattern} {Recognition} ({CVPR})}, 2017.

\bibitem[Sen et~al.(2023)Sen, Chermala, Nagori, Peddigari, Mathur, Prasad, and Jeong]{sen2023shards}
Mrinmoy Sen, Sai~Pradyumna Chermala, Nazrinbanu~Nurmohammad Nagori, Venkat Peddigari, Praful Mathur, BH Prasad, and Moonhwan Jeong.
\newblock Shards: Efficient shadow removal using dual stage network for high-resolution images.
\newblock In \emph{Proceedings of the IEEE/CVF Winter Conference on Applications of Computer Vision (WACV)}, 2023.

\bibitem[Shor and Lischinski(2008)]{shor_2008_shadow}
Yael Shor and Dani Lischinski.
\newblock The shadow meets the mask: Pyramid-based shadow removal.
\newblock \emph{Computer Graphics Forum}, 27\penalty0 (2), 2008.

\bibitem[Sidorov(2019)]{sidorov_2019_conditional}
Oleksii Sidorov.
\newblock Conditional gans for multi-illuminant color constancy: Revolution or yet another approach?
\newblock In \emph{Proceedings of the IEEE/CVF Conference on Computer Vision and Pattern Recognition Workshops (CVPRW)}, 2019.

\bibitem[Simonyan and Zisserman(2015)]{simonyan_2015_vgg}
Karen Simonyan and Andrew Zisserman.
\newblock Very deep convolutional networks for large-scale image recognition.
\newblock In \emph{Proceedings of the International Conference on Learning Representations (ICLR)}, 2015.

\bibitem[Vasluianu et~al.(2021)Vasluianu, Romero, Gool, and Timofte]{Vasluianu_2021_CVPR}
Florin-Alexandru Vasluianu, Andres Romero, Luc~Van Gool, and Radu Timofte.
\newblock Shadow removal with paired and unpaired learning.
\newblock In \emph{Proceedings of the IEEE/CVF Conference on Computer Vision and Pattern Recognition Workshops (CVPRW)}, 2021.

\bibitem[Vicente et~al.(2015)Vicente, Hoai, and Samaras]{vicente_2015_detection}
Tomas F.~Yago Vicente, Minh Hoai, and Dimitris Samaras.
\newblock Leave-one-out kernel optimization for shadow detection.
\newblock In \emph{Proceedings of the IEEE International Conference on Computer Vision (ICCV)}, 2015.

\bibitem[Wan et~al.(2022)Wan, Yin, Wu, Wu, Liu, and Wang]{wan2022style}
Jin Wan, Hui Yin, Zhenyao Wu, Xinyi Wu, Yanting Liu, and Song Wang.
\newblock Style-guided shadow removal.
\newblock In \emph{Proceedings of the European Conference on Computer Vision (ECCV)}, 2022.

\bibitem[Wang et~al.(2020{\natexlab{a}})Wang, Xu, Zhou, Deng, and Yang]{wang_2020_vehicles}
Chunxiang Wang, Hanqing Xu, Zhiyu Zhou, Liuyuan Deng, and Ming Yang.
\newblock Shadow detection and removal for illumination consistency on the road.
\newblock \emph{IEEE Transactions on Intelligent Vehicles}, 5\penalty0 (4), 2020{\natexlab{a}}.

\bibitem[Wang et~al.(2018{\natexlab{a}})Wang, Li, and Yang]{wang_stacked_2018}
Jifeng Wang, Xiang Li, and Jian Yang.
\newblock Stacked {Conditional} {Generative} {Adversarial} {Networks} for {Jointly} {Learning} {Shadow} {Detection} and {Shadow} {Removal}.
\newblock In \emph{Proceedings of the {IEEE}/{CVF} {Conference} on {Computer} {Vision} and {Pattern} {Recognition} (CVPR)}, 2018{\natexlab{a}}.

\bibitem[Wang et~al.(2020{\natexlab{b}})Wang, Hu, Wang, Heng, and Fu]{wang_2020_shadowdet}
Tianyu Wang, Xiaowei Hu, Qiong Wang, Pheng-Ann Heng, and Chi-Wing Fu.
\newblock Instance shadow detection.
\newblock In \emph{Proceedings of the IEEE/CVF Conference on Computer Vision and Pattern Recognition (CVPR)}, 2020{\natexlab{b}}.

\bibitem[Wang et~al.(2021{\natexlab{a}})Wang, Hu, Fu, and Heng]{wang_2021_det}
Tianyu Wang, Xiaowei Hu, Chi-Wing Fu, and Pheng-Ann Heng.
\newblock Single-stage instance shadow detection with bidirectional relation learning.
\newblock In \emph{Proceedings of the IEEE/CVF Conference on Computer Vision and Pattern Recognition (CVPR)}, 2021{\natexlab{a}}.

\bibitem[Wang et~al.(2018{\natexlab{b}})Wang, Liu, Zhu, Tao, Kautz, and Catanzaro]{wang_2018_pix2pixHD}
Ting-Chun Wang, Ming-Yu Liu, Jun-Yan Zhu, Andrew Tao, Jan Kautz, and Bryan Catanzaro.
\newblock High-resolution image synthesis and semantic manipulation with conditional gans.
\newblock In \emph{Proceedings of the IEEE/CVF Conference on Computer Vision and Pattern Recognition (CVPR)}, 2018{\natexlab{b}}.

\bibitem[Wang et~al.(2021{\natexlab{b}})Wang, Yao, Dai, Wang, and Cao]{wang_2021_bmvc}
Xiao Wang, Siyuan Yao, Pengwen Dai, Rui Wang, and Xiaochun Cao.
\newblock Updated paired regions for shadow detection from single image.
\newblock In \emph{Proceedings of the British Machine Vision Conference (BMVC)}, 2021{\natexlab{b}}.

\bibitem[Wang et~al.(2024)Wang, Zhou, Feng, Li, and Li]{wang2024progressive}
Yonghui Wang, Wengang Zhou, Hao Feng, Li Li, and Houqiang Li.
\newblock Progressive recurrent network for shadow removal.
\newblock \emph{Computer Vision and Image Understanding}, 238, 2024.

\bibitem[Wu et~al.(2012)Wu, Zhang, and Vijaya~Kumar]{wu_2012_shadowsdriving}
Qi Wu, Wende Zhang, and B.V.K. Vijaya~Kumar.
\newblock Strong shadow removal via patch-based shadow edge detection.
\newblock In \emph{2012 IEEE International Conference on Robotics and Automation (ICRA)}, 2012.

\bibitem[Yu et~al.(2022)Yu, Zheng, Huang, and Zhao]{yu2023cnsnet}
Qianhao Yu, Naishan Zheng, Jie Huang, and Feng Zhao.
\newblock Cnsnet: A cleanness-navigated-shadow network for shadow removal.
\newblock In \emph{Proceedings of the European Conference on Computer Vision (Workshops) (ECCVW)}, 2022.

\bibitem[Zhang et~al.(2021)Zhang, Martin-Brualla, Kontkanen, and Curless]{zhang_2021_noshadow}
Edward Zhang, Ricardo Martin-Brualla, Janne Kontkanen, and Brian~L Curless.
\newblock No shadow left behind: Removing objects and their shadows using approximate lighting and geometry.
\newblock In \emph{Proceedings of the IEEE/CVF Conference on Computer Vision and Pattern Recognition (CVPR)}, 2021.

\bibitem[Zhang et~al.(2020{\natexlab{a}})Zhang, Long, Zhang, and Xiao]{zhang_2020_risgan}
Ling Zhang, Chengjiang Long, Xiaolong Zhang, and Chunxia Xiao.
\newblock Ris-gan: Explore residual and illumination with generative adversarial networks for shadow removal.
\newblock In \emph{Proceedings of the AAAI Conference on Artificial Intelligence (AAAI)}, 2020{\natexlab{a}}.

\bibitem[Zhang et~al.(1999)Zhang, Tsai, Cryer, and Shah]{zhang_1999_sfs}
Ruo Zhang, Ping-Sing Tsai, J.E. Cryer, and M. Shah.
\newblock Shape-from-shading: a survey.
\newblock \emph{IEEE Transactions on Pattern Analysis and Machine Intelligence}, 21\penalty0 (8), 1999.

\bibitem[Zhang et~al.(2020{\natexlab{b}})Zhang, Barron, Tsai, Pandey, Zhang, Ng, and Jacobs]{zhang_portrait_2020}
Xuaner~Cecilia Zhang, Jonathan~T. Barron, Yun-Ta Tsai, Rohit Pandey, Xiuming Zhang, Ren Ng, and David~E. Jacobs.
\newblock Portrait {Shadow} {Manipulation}.
\newblock \emph{ACM Transactions on Graphics}, 2020{\natexlab{b}}.

\bibitem[Zheng et~al.(2019)Zheng, Qiao, Cao, and Lau]{zhang_2019_det}
Quanlong Zheng, Xiaotian Qiao, Ying Cao, and Rynson~W.H. Lau.
\newblock Distraction-aware shadow detection.
\newblock In \emph{Proceedings of the IEEE/CVF Conference on Computer Vision and Pattern Recognition (CVPR)}, 2019.

\bibitem[Zhong et~al.(2022)Zhong, Lin, You, Zhang, Liu, and Ji]{zhong2022shadow}
Yunshan Zhong, Mingbao Lin, Lizhou You, Yuxin Zhang, Luoqi Liu, and Rongrong Ji.
\newblock Shadow removal by high-quality shadow synthesis.
\newblock \emph{arXiv preprint arXiv:2212.04108}, 2022.

\bibitem[Zhu et~al.(2017)Zhu, Park, Isola, and Efros]{CycleGAN_2017}
Jun-Yan Zhu, Taesung Park, Phillip Isola, and Alexei~A Efros.
\newblock Unpaired image-to-image translation using cycle-consistent adversarial networks.
\newblock In \emph{Proceedings of the IEEE/CVF International Conference on Computer Vision (ICCV)}, 2017.

\end{thebibliography}
}

%\bibliography{egbib}
\end{document}